%% file: main.tex
\documentclass[10pt,twocolumn,letterpaper]{article}
\usepackage{cvpr}\usepackage[T1]{fontenc}\usepackage{hyperref}\def\paperID{0000}\def\confName{CVPR}\def\confYear{2026}
\usepackage[hyphens]{url} % DO NOT CHANGE THIS
\usepackage{graphicx} % DO NOT CHANGE THIS
\usepackage{cvpr}\usepackage[T1]{fontenc}\usepackage{hyperref}\def\paperID{0000}\def\confName{CVPR}\def\confYear{2026}
\usepackage{caption} % DO NOT CHANGE THIS
\usepackage{placeins}
\usepackage{booktabs}
\usepackage{siunitx}
\usepackage{multirow}

\usepackage{amssymb}
\usepackage{amsmath}
\usepackage{makecell}
\usepackage{array}
\usepackage{xcolor}
\usepackage[most]{tcolorbox}
\tcbuselibrary{listings,breakable,skins}
\usepackage{listings}

\lstdefinestyle{promptstyle}{
    basicstyle=\ttfamily\scriptsize,
    breaklines=true,
    breakatwhitespace=false,
    columns=fullflexible,
    keepspaces=true,
    showstringspaces=false,
    upquote=true,
    tabsize=2
}

\newtcblisting{promptlisting}[2][]{
    enhanced,
    breakable,
    listing only,
    listing options={style=promptstyle},
    colback=gray!4,
    colframe=black!65,
    boxrule=0.8pt,
    arc=3pt,
    left=6pt,
    right=6pt,
    top=6pt,
    bottom=6pt,
    title={#2},
    fonttitle=\bfseries,
    #1
}

\newtcolorbox{schemabox}[2][]{
    enhanced,
    breakable,
    colback=gray!4,
    colframe=black!65,
    boxrule=0.8pt,
    arc=3pt,
    left=6pt,
    right=6pt,
    top=5pt,
    bottom=5pt,
    before upper={\raggedright\ttfamily\scriptsize\setlength{\parskip}{0pt}},
    title={#2},
    fonttitle=\bfseries,
    #1
}

\makeatletter\newcommand{\affiliations}[1]{\g@addto@macro\@author{\\#1}}\makeatother

\title{ArtiMo: Agent-Driven Articulated Mesh Animation}
\author{
\mbox{Chunyu Zou\textsuperscript{\rm 1}},
\mbox{Peng Dai\textsuperscript{\rm 1,2,3,\(\dagger\)}},
\mbox{Yi-Hua Huang\textsuperscript{\rm 1}},
\mbox{Ze Yuan\textsuperscript{\rm 1}},
\mbox{Jingwei Huang\textsuperscript{\rm 4}}\\
\mbox{Yeming Yao\textsuperscript{\rm 1}},
\mbox{Xiaojuan Qi\textsuperscript{\rm 1,\(\dagger\)}}
}
\affiliations{
\textsuperscript{\rm 1}The University of Hong Kong \quad
\textsuperscript{\rm 2}University of British Columbia\\
\textsuperscript{\rm 3}Vector Institute \quad
\textsuperscript{\rm 4}Tencent
}

\begin{document}
\maketitle
\begingroup
\renewcommand{\thefootnote}{\fnsymbol{footnote}}
\footnotetext[2]{Corresponding authors.}
\endgroup

\begin{abstract}
Animating articulated 3D meshes via text requires satisfying strict kinematic constraints, modeling causal interactions between parts, and achieving instruction fidelity. Due to the absence of task-specific training data and explicit
articulation supervision, existing data-driven mesh animation methods are largely inapplicable to this setting. To address this, we propose ArtiMo, a novel agent-driven framework for text-guided articulated mesh animation. Operating in a zero-shot manner, ArtiMo develops an agentic pipeline powered by Large Language and Vision-Language Models (LLMs/VLMs) to orchestrate motion generation. By synergizing the explicit kinematic constraints of URDF with the agent's reasoning and planning capabilities, it effectively produces causally coherent part motions and interactions without requiring model fine-tuning. To ensure motion correctness, the agent additionally utilizes a visual self-improvement mechanism: generated animations are rendered into compact keyframes and motion cues, enabling the VLM to iteratively diagnose and correct errors. Furthermore, we contribute a new benchmark dataset spanning 21 articulated object categories, featuring high-quality motion annotations enriched with causal relationships. Extensive experiments demonstrate that ArtiMo significantly outperforms baselines, particularly on complex, causally driven motions. The project page is available at \url{https://zou-2004.github.io/ArtiMo/}.
\end{abstract}

\section{Introduction}
 Animating articulated 3D objects via text is critical for AR/VR, virtual environments, and games. Unlike free-form mesh animation~\cite{jiang2024animate3d, wu2025animateanymesh}, articulated-object animation must respect part-level mechanisms: drawers slide along rails, wheels rotate with body motion, and pressing a pedal may trigger a lid to open. Thus, articulated animation requires not only plausible motion, but also kinematically valid and causally correct part interactions.

Existing text-driven 4D animation methods generate plausible motions for generic meshes by leveraging video diffusion priors~\cite{jiang2024animate3d} or learned motion representations~\cite{wu2025animateanymesh}. However, they typically ignore explicit articulation structures, and therefore struggle with motions governed by joint axes, limits, part dependencies, and causal ordering. Structured asset formats such as the Unified Robot Description Format (URDF)~\cite{quigley2009ros, tola2024understanding} provide links, joints, axes, parent-child relations, and motion limits. However, URDF specifies how parts can move, not how they should move for a given action: it does not tell which part is manipulated, how motion propagates, or whether causal relationships exist.

This reveals a central gap: articulated mesh animation is neither standard text-to-motion generation nor simple URDF-based joint execution. It requires a causal understanding of the articulation: identifying the manipulated part, affected joints, motion propagation, temporal ordering, and causal relationships (e.g., the trash can's pedal vs lid). Addressing this gap requires combining low-level kinematic constraints from URDF with high-level semantic and causal reasoning from VLMs/LLMs.

\begin{figure}
    \centering
    \includegraphics[width=\linewidth]{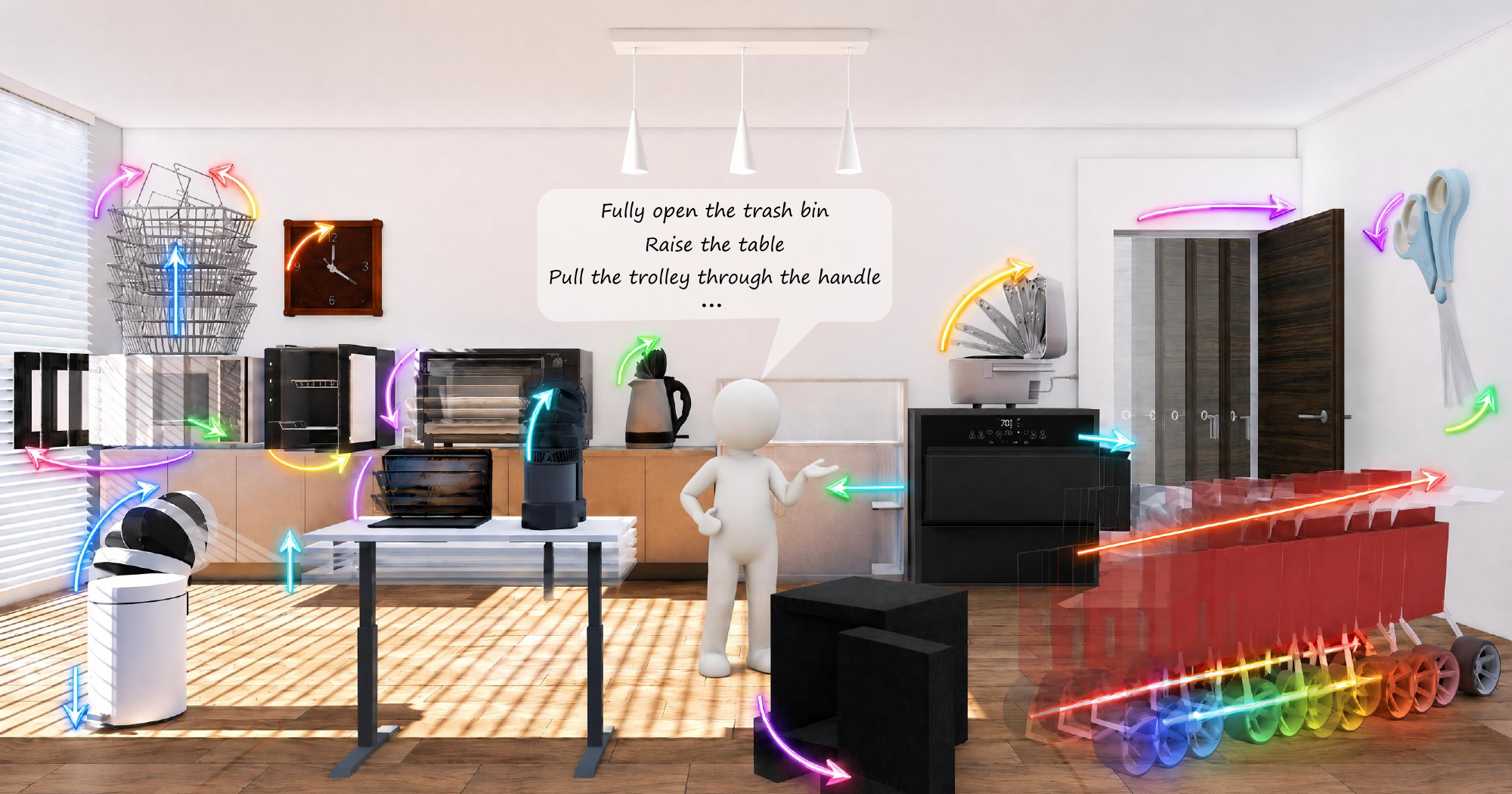}
    \label{fig:teaser}
    \vspace{-0.2in}
    \caption{Given articulated assets and prompts, ArtiMo generates kinematically valid and causally correct motions.}
    \vspace{-0.2in}
\end{figure}

Based on this insight, we propose \textbf{ArtiMo}, a zero-shot agentic framework for text-guided articulated mesh animation. ArtiMo converts each URDF asset into agent-readable representations, including rendered views, colored part overlays, joint summaries, and scale context. It then uses a VLM to ground part semantics and infer action-relevant causal relationships, and an LLM to translate them into executable joint-level trajectories with timing and state transitions. Because all motions are executed under URDF-defined joint constraints, the generated 4D animations remain kinematically valid while following the intended action semantics. Furthermore, ArtiMo introduces a self-improvement critic loop. Instead of relying on single-pass planning, it renders the generated 4D animation into a compact representation (keyframes with motion cues), allowing a VLM critic to efficiently diagnose errors, such as wrong direction, incorrect magnitude, or unintended motion. This diagnostic feedback is used to revise and enhance the motion plan, progressively improving motion correctness.

To address the lack of evaluation benchmarks for articulated mesh animation, we introduce a new benchmark for text-conditioned articulated mesh animation, comprising 225 annotated motion sequences across 21 object categories with both causal and non-causal actions. Experiments show that ArtiMo substantially outperforms existing 2D/4D animation baselines, while ablations confirm the importance of structured agentic planning and the visual critic.

Our main contributions are summarized as follows:
\begin{itemize}
     \item We introduce the task of text-driven articulated mesh animation, which aims to generate 4D animations that satisfy both URDF-defined kinematic constraints and causal relationships among object parts. 
     
     \item We propose ArtiMo, a zero-shot agentic framework that integrates VLM-based visual grounding and causal reasoning, LLM-based motion planning, URDF-constrained motion execution, and a visual self-improvement loop that leverages rendered keyframes and motion cues to iteratively improve motion correctness.

     \item We construct a new benchmark comprising 225 articulated motion annotations across 21 object categories. Experiments demonstrate that ArtiMo consistently outperforms 2D and 4D animation baselines, particularly on kinematically coupled and causally dependent motions.
 \end{itemize}

\section{Related Work}

\noindent\textbf{Object animation and interaction.}
Recent works animate meshes or images using learned motion and video priors. Animate3D~\cite{jiang2024animate3d} transfers multi-view video diffusion priors to static 3D assets, MotionDreamer~\cite{uzolas2025motiondreamer} uses semantic video features for zero-shot mesh animation, AnimateAnyMesh~\cite{wu2025animateanymesh} generates text-conditioned motion for arbitrary mesh topology, and Puppet-Master~\cite{li2025puppetmaster} produces part-level image animation from sparse controls. These methods primarily optimize visually plausible motion without explicitly modeling object-specific kinematic structures. A parallel line of work recovers such structures from geometry, images, or observed motion, including movable parts, joint types, axes, limits, and kinematic hierarchies~\cite{wang2019shape2motion,jiang2022ditto,liu2023paris,lei2023nap,chen2024urdformer,le2025articulate,li2025particulate}. Building on articulated representations, interaction methods predict actionable parts, contact locations, motion directions, or manipulation trajectories~\cite{geng2023gapartnet,mo2021where2act,eisner2022flowbot3d,wu2022vatmart}. Classical procedural animation, physics simulation, and robotic task-and-motion planning further realize object motion through authored rules, physical controllers, or actions of an external manipulator~\cite{witkin1988spacetime,perlin1995responsive,todorov2012mujoco,xiang2020sapien,garrett2021tamp,chen2025robotwin2}. Unlike previous methods, ArtiMo explicitly takes the target object's URDF as input and translates an open-vocabulary action instruction into an executable, kinematically valid, and causally ordered joint-state animation, without being restricted to a specific object category.
\vspace{0.1in}

\noindent\textbf{Agentic graphics workflows and self-refinement.}
Agentic workflows organize language models into role-specific stages for planning, execution, evaluation, and revision. Self-refinement methods further improve intermediate outputs through repeated critique and correction without additional training~\cite{madaan2023selfrefine}. In graphics, LayoutGPT~\cite{feng2023layoutgpt} and Holodeck~\cite{yang2024holodeck} use language models for structured spatial planning and environment generation. SAGE~\cite{xia2026sage} extends this process with graphics tools and critic-guided scene refinement. More recent systems, including BlenderGym~\cite{gu2025blendergym}, IR3D-Bench~\cite{liu2025ir3dbench}, and VIGA~\cite{yin2026viga}, ground graphics agents in program execution and rendered feedback, enabling generated scenes or programs to be evaluated and revised in a closed loop.
In contrast, ArtiMo specializes agentic workflows for articulated-object motion by grounding language instructions to URDF links and generating executable, time-parameterized joint plans.

\begin{figure*}[t]
    \centering

    \includegraphics[width=\textwidth]{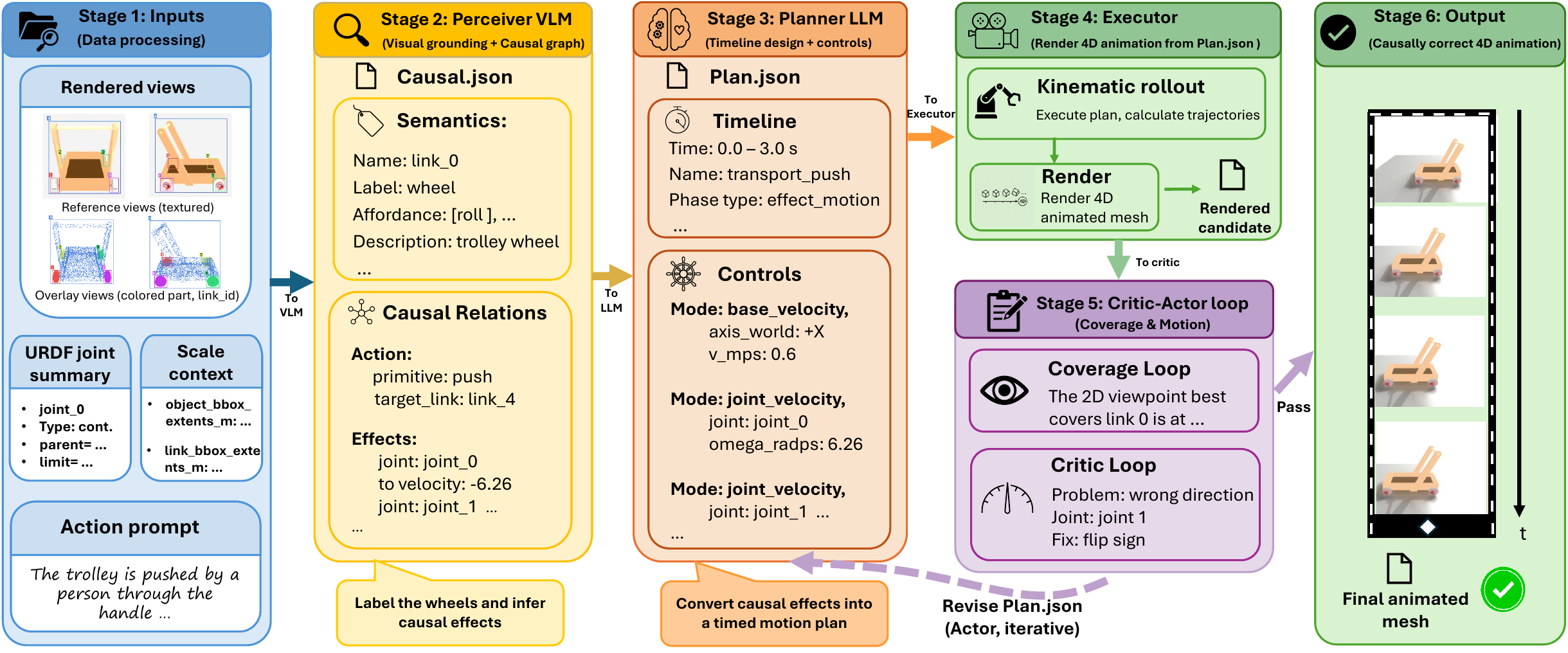}
    \caption{Overview.
Given a preprocessed articulated mesh, its URDF, and an action prompt, our system performs link analysis, causal/motion planning, and
critic--actor refinement to generate a kinematically and causally
correct 4D animation.}
\vspace{-0.1in}
    \label{method}

\end{figure*}

\section{Method}
We introduce \textit{ArtiMo}, an agentic system that takes an articulated mesh with its URDF description and an action prompt as inputs, and generates a GL Transmission Format (GLTF) animation that executes the specified
action while maintaining kinematic and causal correctness, as
illustrated in Fig.~\ref{method}. We first formulate the problem in
Sec.~\ref{sec:pd}. We then present the proposed framework in
Sec.~\ref{agent_driven_animation_generation}, including the
single-pass agentic animation pipeline and
critic--actor refinement loop. Finally, Sec.~\ref{benchmark}
introduces the benchmark construction and evaluation metrics.

\subsection{Problem Definition}
\label{sec:pd}
We explore the problem of generating causally correct 4D animations from structured input files. Specifically, we assume as input a URDF file $\mathcal{U}$ describing the articulation of an object, which defines a set of links (rigid parts) $\mathcal{L} = \{l_1, \dots, l_N\}$, joints connecting the links $\mathcal{J} = \{j_1, \dots, j_M\}$, and motion constraints such as parent-child relationships, joint types, motion axes, joint origins, and limits (see supplementary Sec. 1.1 for details). Notably, URDF does not encode causal relationships or time-varying motions. In addition, we are given mesh files $\mathcal{M}$ representing the geometric shapes of the links, and an action prompt $\mathcal{A}$ specifying a desired behavior to execute.

Our goal is to generate the object's 4D animation in GLTF, which encodes the generated action-conditioned motion as executable animation channels over a discrete timeline $\mathcal{X} = \{\mathbf{s}_1, \dots, \mathbf{s}_T\}$. At each time step $t$, $\mathbf{s}_t = \{s_t^{(j_1)}, \dots, s_t^{(j_M)}\}$ denotes the configurations of all joints in the articulated object. The resulting animation must satisfy both the kinematic constraints specified by the URDF and the causal requirements implied by the action prompt.

We categorize the articulated mesh animation into four types: \textit{(1) Independent motion.} The action prompt is directly mapped to a joint motion without coupled effects or temporal dependencies, e.g., pulling a drawer. \textit{(2) Mechanically coupled motion.} Motion of a target link induces corresponding motions in other mechanically coupled links; for example, pressing the door-release button on a microwave causes the microwave door to open accordingly.\enspace\textit{(3) State-dependent motion.} Certain motions require prerequisite states before subsequent actions can occur; for example, pulling out an oven tray first requires opening the oven door. \textit{ (4) Environment-mediated motion.} Motion is influenced by both the action prompt and environmental factors. For example, pushing a toy car requires coordinated body translation and wheel rotation consistent with rolling.

Given the complexity of articulated mesh animation, we investigate leveraging the reasoning capabilities of VLMs and LLMs to infer causal relationships and plan executable motions by developing an agentic system that generates the animation sequence $\mathcal{X}$:
\begin{equation}
    \mathcal{X} = \text{Agent}(\mathcal{A}, \mathcal{U}, \mathcal{M}).
    \label{equ:overall}
\end{equation}

\subsection{Agent-Driven Animation Generation}
\label{agent_driven_animation_generation}
As shown in Fig.~\ref{method}, our agentic system generates a time series of joint states from the input URDF $\mathcal{U}$, meshes $\mathcal{M}$, and action prompt $\mathcal{A}$. At the core of our framework is a single-pass agentic pipeline for  articulated mesh animation generation and a self-improvement loop that iteratively corrects previously generated erroneous motions.

\subsubsection{Single-pass agentic animation system.}
\label{single_pass}
In this stage, the system first preprocesses the input data to make it suitable for agent-based processing (Fig.~\ref{method} stage 1). It then invokes the perceiver (Fig.~\ref{method} stage 2), planner (Fig.~\ref{method} stage 3), and executor (Fig.~\ref{method} stage 4) to analyze the visual content, reason about causal relationships, generate the motion plan, and render the resulting 4D animations.

\textit{Data preprocessing.}
Our system transforms raw URDF $\mathcal{U}$ and mesh $\mathcal{M}$ into structured visual inputs suitable for VLM processing, including: \textit{(1) Reference images.} The textured mesh is rendered from four viewpoints (front, back, left, and right) to provide the agent with the object’s appearance and geometry. \textit{(2) Overlay images.} Based on the URDF-defined links/segments $\mathcal{L}$, we assign each segment a random color and render overlay images from the same camera viewpoints. To reveal interior or occluded structures (e.g., trays inside a closed microwave), we render sampled surface points instead of watertight meshes, allowing internal components to remain partially visible through the resulting gaps. 
Such overlay images provide the agent with strong visual distinctions. Additionally, we overlay bounding boxes on images to further improve link localization and identification (see Supplementary Fig.~1).
\textit{(3) Joint summary.} We simplify the URDF into a concise joint summary $\mathcal{J}$, including motion type, parent-child relationship, axis, origin, and limit. It is designed to be short but effective. \textit{(4) Scale context.} Extracted from the mesh file to define the scale of the object. This enables the agent to interpret quantitative expressions in the action prompt and estimate appropriate joint motion magnitudes. (See Supplementary Sec.~1.3 for details.)

\begin{figure}
    \centering
    \includegraphics[width=\linewidth]{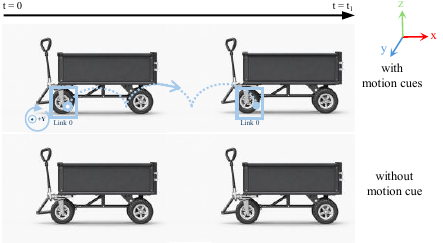}
    \vspace{-0.2in}
    \caption{
 Motion-cue representation for visual criticism. For each target link, we augment the start and end frames with motion trajectories and explicit translational and rotational motion cues (top). Without these cues, the rendered RGB frames provide limited visual evidence of the wheel's motion, making accurate motion assessment challenging (bottom).
}
    \label{fig:motion_cue}
    \vspace{-0.1in}
\end{figure}

\textit{Perceiver.}
Once the aforementioned metadata is prepared, the agent invokes the VLM to analyze each link and infer causal relationships, guided by Chain-of-Thought (CoT)-style prompt designs. First, the VLM is prompted to describe the name, affordance, and description of each link. This improves the VLM’s understanding of the input by encouraging detailed and structured descriptions of each component. Second, the VLM is prompted to describe joint motions and their causal relationships. Specifically, given the action prompt $\mathcal{A}$, it identifies the directly manipulated target link and infers the corresponding joint motion (e.g., translation distance, rotation angle, etc.) induced by the action prompt. Moreover, it also analyzes and specifies the motions of other joints that are causally affected by the movement of the target link (e.g., door-release button $\rightarrow$ door). The output of \textit{Perceiver} is a \texttt{Causal.json} file, containing coarse descriptions of motion and causal relationships. (See Supplementary Sec. 2.1 for prompt designs.)

\vspace{0.1in}
\textit{Planner.}
We employ an LLM as the animation planner to generate a fine-grained temporal sequence of joint configurations $\mathcal{X}$, conditioned on the \texttt{Causal.json} file (i.e., coarse motions and causal relationships) and the processed object metadata. Specifically, the LLM predicts the overall animation duration, orchestrates fine-grained key timestamps, and determines the state transitions of each joint. For motions not triggered simultaneously (e.g., state-dependent motions in Sec.~\ref{sec:pd}), it further infers appropriate latencies and temporal offsets to correctly order dependent animations, ensuring causal consistency throughout the animation. The resulting motion plan is saved as a  \texttt{Plan.json} file. (See Supplementary Sec. 2.2 for prompt designs.)

\vspace{0.1in}
\textit{Executor.}
Given \texttt{Plan.json}, the executor interprets the planned joint configurations under the URDF-defined articulation, interpolates between scheduled key states to construct continuous joint trajectories, and propagates link transforms through the parent--child kinematic hierarchy. Blender then exports the resulting motion as a standard GLTF animation.

\subsubsection{Critic--actor loop for self-improvement.}
\label{loop}

Animations generated via single-pass inference are not sufficiently robust and may exhibit errors, such as motion direction (e.g., a link moves in the wrong direction due to an incorrect velocity sign), magnitude (e.g., motions that are excessively small or large), and return behavior (e.g., a pressed button failing to return to its original state). To address these issues, we design a self-improvement loop that evaluates rendered 2D animations and generates correction instructions to refine erroneous motion predictions (Fig.~\ref{method}, stage 5).

\begin{figure*}[t]
    \centering
    \includegraphics[width=\linewidth]{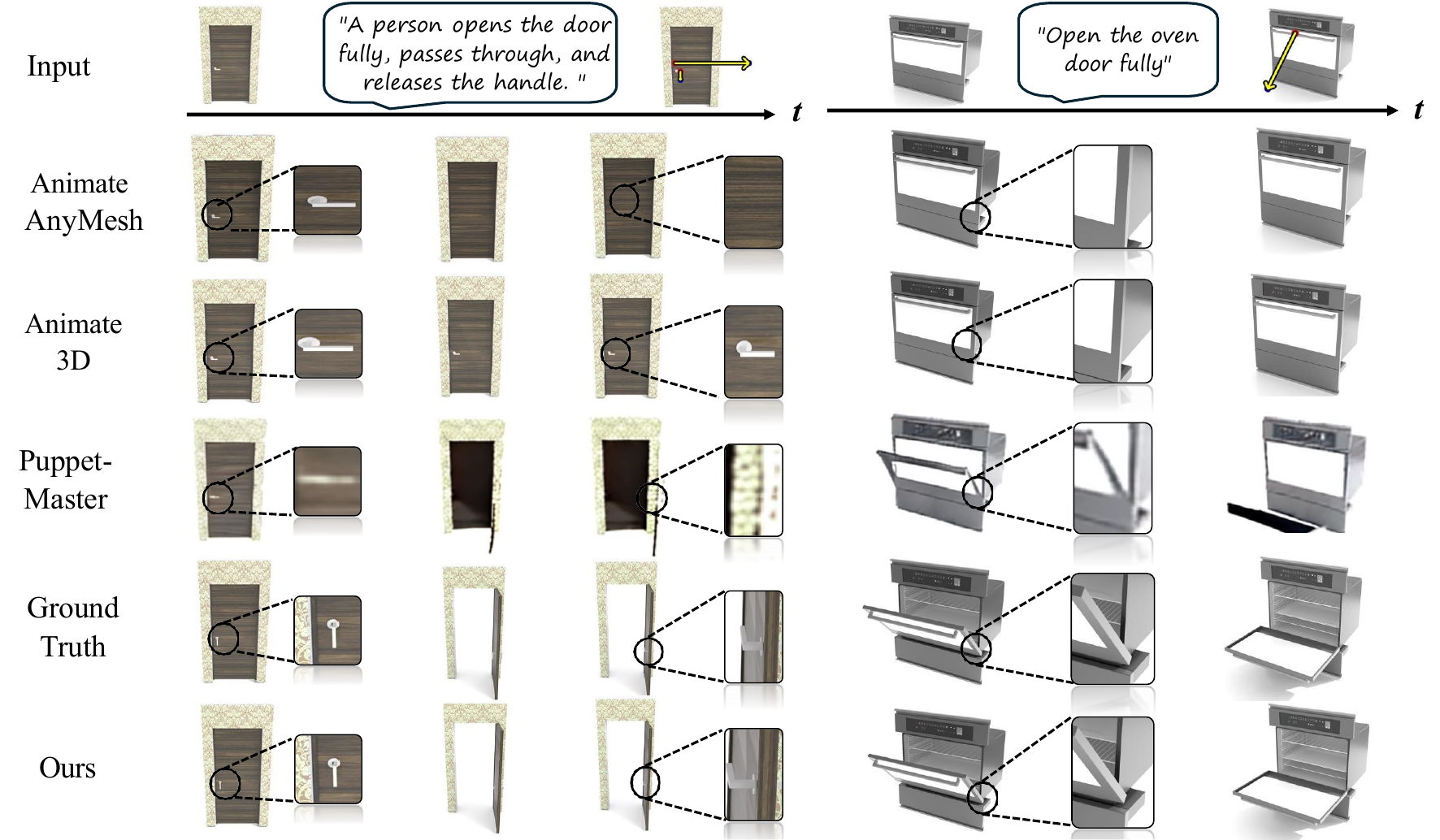}

    \caption{Qualitative comparison with state-of-the-art methods. Compared with existing animation methods, our approach better preserves articulated structure and produces action-consistent joint-level motion.}
    \vspace{-0.1in}
    \label{qualitative_comparison}

\end{figure*}

\vspace{0.1in}
\textit{Critic.} To evaluate the visual correctness of the generated articulated mesh animations, we sample multiple camera viewpoints and select the one that best covers the link motions, then render the 4D animations into 2D videos for VLM-based assessment. Specifically, we first sample four views (front, back, left, and right) and select the one providing the highest link visibility and motion coverage. If none of these views is satisfactory, the VLM proposes alternative viewpoints for 2D video rendering. See Supplementary Sec.~2.3 for detailed Coverage loop designs and VLM output schema.

The key challenge is how to efficiently and effectively present 2D videos to the visual critic. Using too many frames incurs substantial computational overhead, whereas sparse sampling may miss critical motions. Moreover, as illustrated in Fig.~\ref{fig:motion_cue} bottom, motions are often difficult to perceive directly from RGB frames due to their limited visual saliency, making accurate motion assessment challenging.

To address this challenge, we convert the rendered 2D videos into a compact representation consisting of key frames and explicit motion cues that enhance the visibility of object motion, as illustrated in Fig. ~\ref{fig:motion_cue} top. Specifically, for each moving link, we sample a point on its surface and track its position over time, forming a global motion trajectory that describes the link's motion patterns. Additionally, we combine straight or circular arrows with a cross-in/dot-out sign to explicitly indicate the global translational and rotational motion directions of the link. 
Based on the compact motion representation, the VLM critic performs
structured analysis of the observed link motion and identifies local
errors from a pre-defined error pool. It then
outputs a bounded repair hint for the detected error; detailed motion cue representations, diagnostic fields and output definitions are provided in Supplementary Sec.~2.4.

\vspace{0.1in}
\textit{Actor.} Given the structured diagnosis and repair hint produced by the visual critic, the actor deterministically refines the animation by updating the previous motion plan \texttt{Plan.json}. 
The revised plan is then re-executed and rendered again into the compact video representation for further error diagnosis. This iterative process continues until the visual critic detects no remaining errors or the maximum number of iterations is reached. Through this loop-based refinement, we ultimately obtain a corrected and visually improved animation (Fig.~\ref{method} Output).

\subsection{Benchmark Construction}
\label{benchmark}
\label{dataset}

Existing articulated-object datasets ~\cite{xiang2020sapien, jin2026artvip,li2025particulate,lightwheel_simready_2025} primarily provide object geometry and joint structures, but they lack text-conditioned motion ground truth, making them unsuitable for evaluating articulated mesh animation generation. To address this limitation, we construct a new benchmark for this task, including an articulated-mesh animation dataset and corresponding evaluation metrics.

\begin{figure*}[t]
\centering

\includegraphics[width=\textwidth]{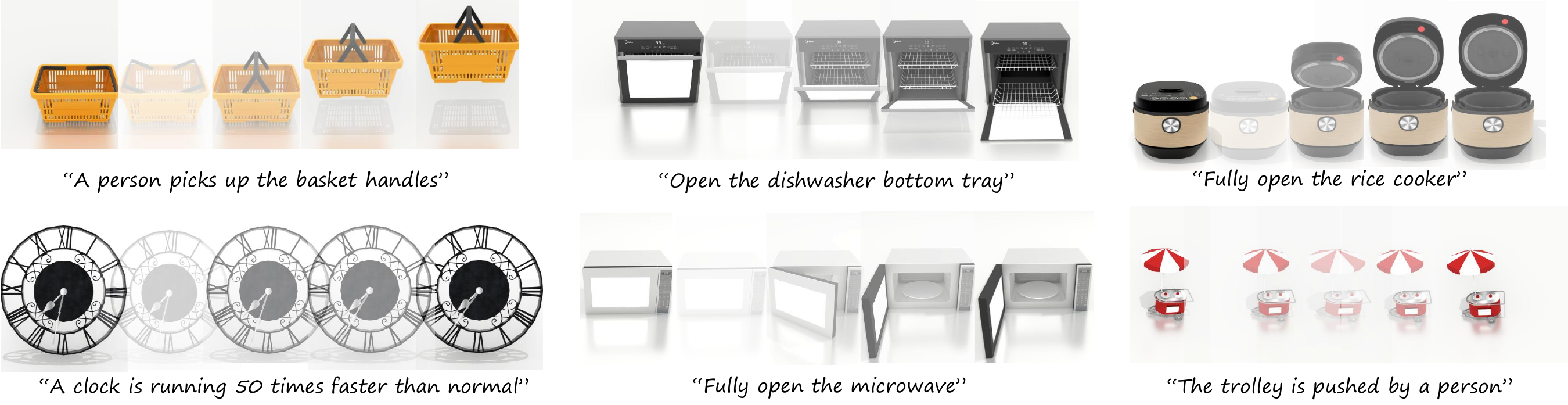}
\vspace{-0.2in}
\caption{Examples of generated articulated mesh animations.
Our method generates action-conditioned 4D animations for diverse articulated objects.}
    \label{result}
    \label{qualitative result}
    \vspace{-0.15in}

\end{figure*}

\paragraph{Dataset construction.}
We collect articulated assets with URDF annotations from
PartNet-Mobility~\cite{xiang2020sapien},
ARTVIP~\cite{jin2026artvip}, and
LightWheel~\cite{lightwheel_simready_2025}. The benchmark is expert-curated
following a standardized annotation protocol, in which annotators define
and verify the manipulated and affected links, joint motions, temporal
relations, and release--return behavior under the URDF constraints.
The structured motion plans are iteratively edited and rendered until
the resulting animations correctly satisfy the action instruction,
causal relations, and common-sense kinematic behavior. Each benchmark
annotation is stored as a structured record of ordered motion phases,
per-phase joint and link constraints, and temporal relations, together
with the corresponding verified animation. The final benchmark contains 225 animations across 21 object
categories; further details on its distribution and format are
provided in Supplementary Sec.~3.

\paragraph{Evaluation metrics.}
We segment the ground-truth animation $\mathcal{X}^{gt}$ into $K$ distinct motion phases based on the stopping timestamp of each link motion, and construct a set of key states $(\mathcal{X}^{gt}_{1}, \dots, \mathcal{X}^{gt}_K)$ by selecting the endpoint of each motion phase for metric evaluation.
To accommodate differences in animation duration and motion speed between the ground-truth and predicted animations,
we align ground-truth key states with an ordered sequence of states in the
predicted animation. Specifically, each ground-truth key state
$\mathcal{X}^{gt}_k$ is matched to the visually closest predicted
state $\mathcal{X}^{pred}_{\tau_k}$, subject to
$\tau_1 < \tau_2 < \cdots < \tau_K$. We then denote the matched
state as $\mathcal{X}^{pred}_k =
\mathcal{X}^{pred}_{\tau_k}$. This order-preserving alignment allows
the exact timing of each phase to vary while retaining the states' order. Given the resulting $K$ matched state pairs, we compute part-aware
scores to prevent small but action-critical parts, such as buttons,
from being dominated by large components. We average the similarity scores of both the dynamic and static parts over all matched $K$ state
pairs:

{\small
\begin{equation}
\begin{aligned}
S^{3D}
=
\frac{1}{K}\sum_{k=1}^{K}
\Bigg[
&\lambda \frac{1}{|\mathcal{D}|}
\sum_{p \in \mathcal{D}}
s^{3D}(\mathcal{X}_{k,p}^{gt},\mathcal{X}_{k,p}^{pred})
\\
+{}&
(1-\lambda)\frac{1}{|\mathcal{S}|}
\sum_{p \in \mathcal{S}}
s^{3D}(\mathcal{X}_{k,p}^{gt},\mathcal{X}_{k,p}^{pred})
\Bigg].
\end{aligned}
\end{equation}
}

Here, $s^{3D}$ denotes a specific 3D similarity metric,
$\mathcal{D}$ and $\mathcal{S}$ denote the sets of dynamic and static
parts, respectively, and $\lambda$ controls their relative
contributions. 
We set $\lambda=0.8$ to emphasize the
action-relevant dynamic parts while retaining the contribution of
static geometry.
In this paper, we employ three 3D metrics: (1) Part-wise generalized Intersection over Union~\cite{rezatofighi2019giou} (\textbf{$\text{P\_{gIoU}}$}) measures coarse per-part 3D spatial overlap using axis-aligned bounding boxes; (2) Part-wise Point Consistency~\cite{fan2017pointset} (\textbf{$\text{P\_{PC}}$}) measures fine-grained per-part surface alignment based on sampled surface points; and (3) Part-wise Occupancy Consistency~\cite{mescheder2019occupancy} (\textbf{$\text{P\_{OccF1}}$}) measures per-part volumetric occupancy agreement after voxelization. Similarly, for the 2D variants used when comparing with image-based animation methods, we render the 3D object into 2D mask images and compute the following three 2D metrics: (1) \textbf{$\text{P\_{MaskIoU}}$}  measures per-part mask overlap in the rendered image space~\cite{lin2014coco}; (2) \textbf{$\text{P\_{ContourCD}}$} measures per-part contour distance between predicted and ground-truth masks~\cite{borgefors1988chamfer}; and (3) \textbf{$\text{P\_{BoundaryF1}}$} measures per-part silhouette boundary alignment~\cite{arbelaez2011contour}.

\section{Experiments}
\subsection{Setup}

\textit{ArtiMo} integrates large foundation models for perception, reasoning, and planning. We use Gemini 3.1 Pro as the VLM and ChatGPT 5.4 as the LLM, and use Blender to execute the motion plans and render the generated 4D articulated mesh animations into 2D video frames. VLM and LLM inference is performed through their respective APIs. All local motion execution, rendering, and 3D geometry processing are
performed on a workstation equipped with a single NVIDIA GeForce RTX
4090 GPU; Blender is used for execution and rendering, and PyTorch3D
is used for GPU-accelerated geometry processing. We allow at most three critic--actor refinement rounds after the initial execution. The coverage module permits at most two alternative-view proposals after evaluating the four canonical views.

\subsection{Baselines}

We compare against representative 2D and 3D animation generation methods. AnimateAnyMesh~\cite{wu2025animateanymesh} and Animate3D~\cite{jiang2024animate3d} operate on 3D meshes, while Puppet-Master~\cite{li2025puppetmaster} animates 2D images using drag interactions. Since these baselines do not assume oracle articulation input, we additionally evaluate \textit{ArtiMo} with URDFs predicted by Particulate~\cite{li2025particulate}, a recent 3D object articulation method. This setting compares methods under similar input conditions, while the oracle-URDF setting reflects the upper-bound performance of our framework when accurate articulation structure is available.

\subsection{Results}

Our proposed \textit{ArtiMo} enables text-driven 4D animation generation for generic articulated objects, producing mesh sequences with improved action following, causal correctness, and geometry consistency. Representative examples are shown in Fig.~\ref{result}.

\setcounter{figure}{6}\begin{figure*}[!t]\centering\includegraphics[width=\textwidth]{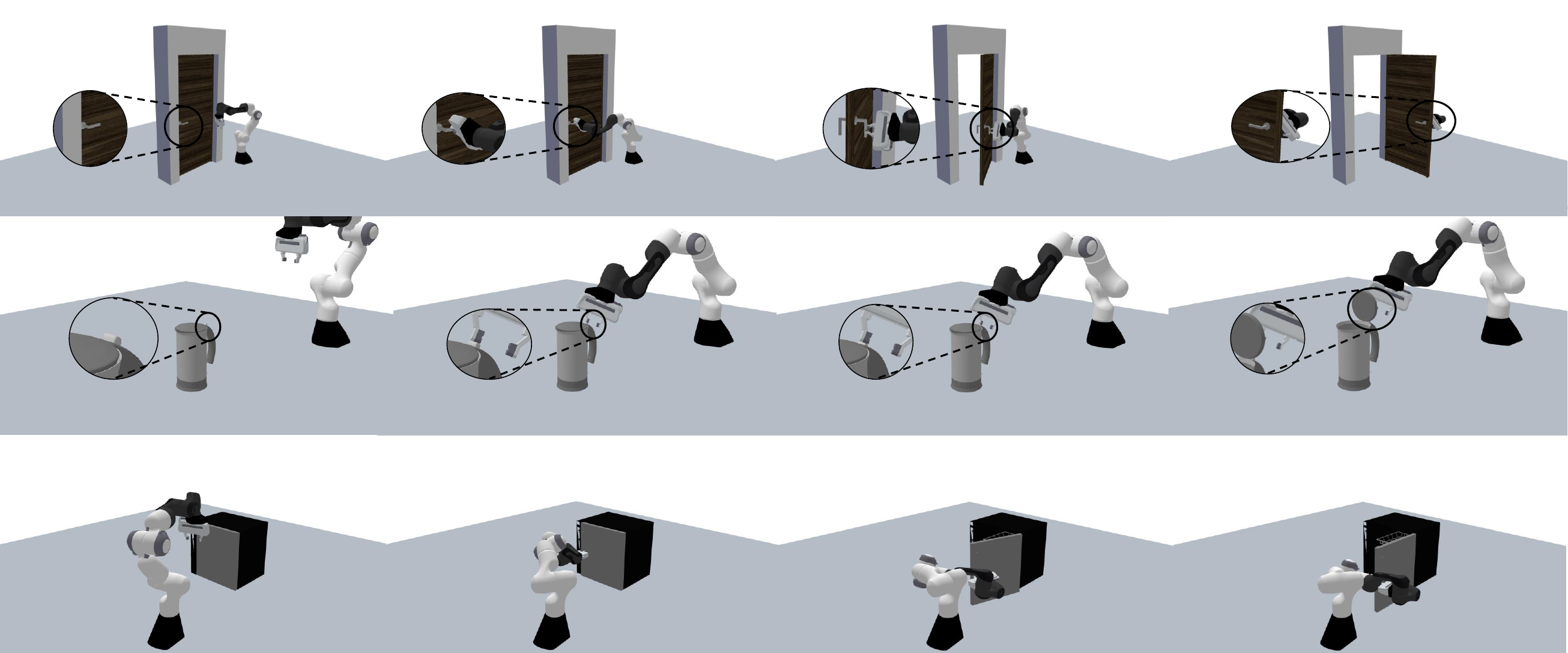}\caption{\textbf{From articulated motion understanding to robotic manipulation.} ArtiMo-generated motions encode not only object-link trajectories but also the interactions and causal dependencies that produce them. For the door, it first rotates the handle before opening the door, respecting the causal dependency between the two articulated stages. For the kettle, the robot presses the release button instead of directly manipulating the lid, allowing the internal mechanism to trigger the generated lid-opening motion. For the dishwasher, the robot grasps the handle and follows the generated door motion. These examples illustrate how ArtiMo provides object-centric interaction targets that can be retargeted to downstream robotic execution.}\label{application}\vspace{-0.15in}\end{figure*}\setcounter{figure}{5}\paragraph{Qualitative comparison.}
As illustrated in Fig.~\ref{qualitative_comparison}, existing methods exhibit clear limitations in action-conditioned articulated animation. AnimateAnyMesh~\cite{wu2025animateanymesh} is designed for text-driven animation of arbitrary meshes and can often preserve the overall object shape. However, it does not explicitly model link-level articulation or causal dependencies, and therefore its generated animations often remain nearly static or fail to follow the required action over time. Animate3D~\cite{jiang2024animate3d} relies on multi-view video diffusion to generate dynamic 3D content. While it can produce apparent motion, it often favors global object deformation or overall movement rather than the intended joint-level mechanism. This makes it less reliable for actions requiring precise hinge rotation, release--return behavior, or control-to-effect motion. During evaluation, its best-matched frames are often static or early frames before dramatic global movement occurs, resulting in weak articulated action execution. Puppet-Master~\cite{li2025puppetmaster} operates in the 2D image domain with drag-based controls. Although it can create plausible image-space motion, the drag input is ambiguous for specifying 3D joint axes, target links, hidden parts, and temporal ordering. This limitation is especially problematic for articulated objects with occluded components or coupled mechanisms, such as internal trays, handles, buttons, or parts behind the visible surface. As a result, Puppet-Master can produce visually plausible 2D deformations but often fails to recover the correct 3D part interaction or causal motion sequence. In contrast, ArtiMo grounds animation generation in URDF articulation, action semantics, and phase-level causal planning. Our method better preserves object geometry while executing the intended action, including hinge-axis motion, small control-part movements, link-to-link causal responses, and release--return behavior. More results are provided in Supplementary Fig.~2.

\paragraph{Quantitative comparison.}
We evaluate ArtiMo and baselines on our benchmark, which contains both causal and non-causal articulated motions, and report results in Table~\ref{3D_eval} and Table~\ref{2D_eval}.
Across all metrics, ArtiMo achieves the best performance, showing that explicit articulation-aware planning leads to more accurate action execution. Notably, even when using articulation predicted by Particulate rather
than the oracle URDF, ArtiMo achieves a P\_gIoU of 0.450, substantially
outperforming generic mesh animation methods such as AnimateAnyMesh
(0.019) and Animate3D ($-0.130$). This result shows that the advantage
of our framework persists under a mesh-only setting and does not rely
solely on access to ground-truth articulation.
AnimateAnyMesh~\cite{wu2025animateanymesh} and Animate3D~\cite{jiang2024animate3d} perform worse because they mainly target generic mesh animation rather than causal articulated motion. Puppet-Master~\cite{li2025puppetmaster} is evaluated only in the 2D setting and remains limited by its image-space drag interface. Overall, the quantitative results confirm that ArtiMo better preserves geometry and consistently aligns with reference motion annotations. We provide the additional experiments, including per-motion-type breakdown, critic diagnostic accuracy, runtime and model calls, and multiple-run variance in Supplementary Sec. 4.

\begin{figure}
    \centering
\includegraphics[width=\linewidth]{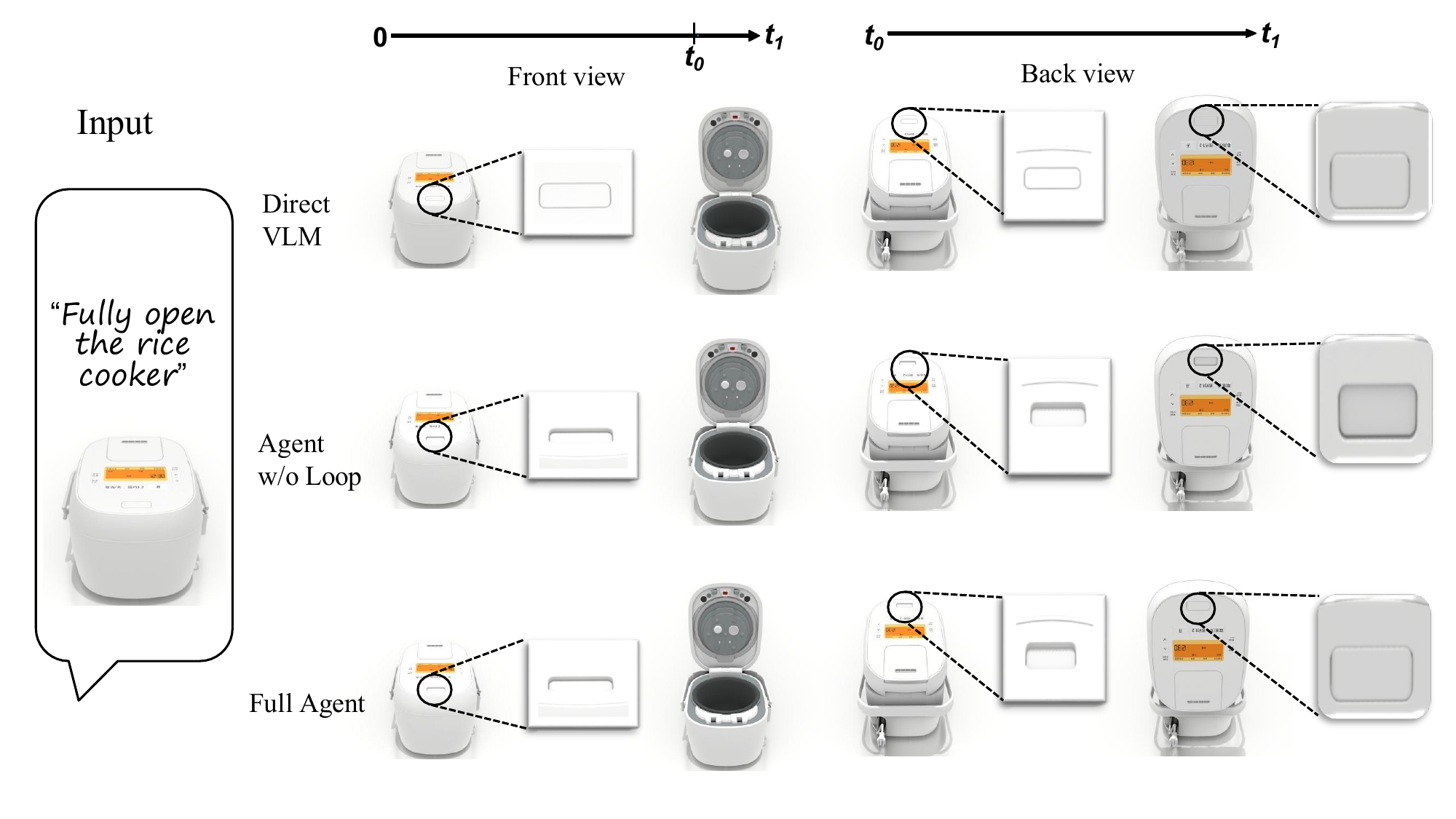}

    \caption{Qualitative ablation of the refinement loop.
The full agent better captures both the main articulated motion and subtle causal behaviors, such as button release--return.}
\vspace{-0.15in}
    \label{ablation_image}
\end{figure}

\subsection{Ablation Study}

We conduct ablation studies to validate the key components of our agentic animation framework. We compare three variants: \textit{Direct VLM}, which directly generates an animation plan without structured agentic modules; \textit{Agent w/o Loop}, which retains single-pass agentic planning but removes iterative critic--actor refinement; and \textit{Full Agent}, which includes both structured planning and loop-based correction. Quantitative results are shown in Table~\ref{Ablation}, and qualitative examples are shown in Fig.~\ref{ablation_image}. The single-pass agentic pipeline substantially improves P\_gIoU from
0.703 to 0.962 and P\_OccF1 from 0.636 to 0.848, demonstrating the
importance of structured perception, causal reasoning, and temporal
motion planning. Incorporating the refinement loop further increases
these scores to 0.985 and 0.899, respectively, showing that iterative
visual correction provides additional gains beyond the initial plan.

The qualitative ablation also demonstrates the benefit of our module designs and iterative refinement. In the rice cooker example shown in Fig.~\ref{ablation_image}, the action requires not only the lid to open, but also the button to exhibit correct self-return behavior after actuation. The Direct VLM produces insufficient lid motion and fails to recover the button's release--return behavior. The \textit{Agent w/o Loop} improves the primary motion through explicit phase decomposition and joint-level planning, yet still misses subtle behaviors such as incomplete self-return. By contrast, \textit{Ours (Full)} employs a critic--actor loop that evaluates rendered intermediate results and iteratively refines the motion plan. This process successfully corrects the remaining errors, resulting in more causally consistent and plausible 4D animations.

\begin{table}[t]
\centering
\caption{3D evaluation. All metrics are part-normalized; higher is better.}
\label{3D_eval}

\resizebox{.8\columnwidth}{!}{
\begin{tabular}{
    l
    S[table-format=-1.3]
    S[table-format=1.3]
    S[table-format=1.3]
}
\toprule
Method & {P\_gIoU} & {P\_PC} & {P\_OccF1} \\
\midrule
Ours (Full) & {\textbf{0.985}} & {\textbf{0.965}} & {\textbf{0.899}} \\
Ours + Particulate & {\underline{0.450}} & {\underline{0.446}} & {\underline{0.170}} \\
AnimateAnyMesh & 0.019 & 0.251 & 0.118 \\
Animate3D & -0.130 & 0.185 & 0.076 \\
\bottomrule
\end{tabular}
}
\end{table}

\begin{table}[t]
\centering
\caption{2D part-wise evaluation. $\uparrow$ higher is better; $\downarrow$ lower is better.}

\label{2D_eval}
\setlength{\tabcolsep}{3.5pt}
\resizebox{\columnwidth}{!}{
\begin{tabular}{
    l
    S[table-format=1.3]
    S[table-format=1.3]
    S[table-format=1.4]
}
\toprule
Method
& {P\_MaskIoU $\uparrow$}
& {P\_BoundaryF1 $\uparrow$}
& {P\_ContourCD $\downarrow$} \\
\midrule
Ours (Full) & \textbf{0.731} & \textbf{0.970} & \textbf{0.0095} \\
Ours + Particulate & \underline{0.530} & \underline{0.745} & \underline{0.0530} \\
AnimateAnyMesh & 0.355 & 0.598 & 0.1600 \\
Animate3D & 0.267 & 0.503 & 0.1870 \\
Puppet-Master & 0.290 & 0.525 & 0.2380 \\
\bottomrule
\end{tabular}
}

\end{table}

\begin{table}[!t]
\centering
\caption{Ablation study of our method. All metrics are part-normalized and higher is better.}
\label{Ablation}
\setlength{\tabcolsep}{3.5pt}

\resizebox{.75\columnwidth}{!}{
\begin{tabular}{
    p{2.4cm}
    S[table-format=1.3]
    S[table-format=1.3]
    S[table-format=1.3]
}
\toprule
Method & {P\_gIoU} & {P\_PC} & {P\_OccF1} \\
\midrule
Direct VLM      & 0.703 & 0.728 & 0.636 \\
Agent w/o Loop  & \underline{0.962} & \underline{0.936} & \underline{0.848} \\
Ours (Full)     & \textbf{0.985} & \textbf{0.965} & \textbf{0.899} \\
\bottomrule
\end{tabular}
}
\end{table}

\iffalse

\begin{figure*}[t]
\centering

\includegraphics[width=\textwidth]{images/application_figure.pdf}
\vspace{-0.2in}
 \caption{
    \textbf{From articulated motion understanding to robotic manipulation.}
    ArtiMo-generated motions encode not only object-link trajectories but also the interactions and causal dependencies that produce them.
    For the door, it first rotates the handle before opening the door, respecting the causal dependency between the two articulated stages.
    For the kettle, the robot presses the release button instead of directly manipulating the lid, allowing the internal mechanism to trigger the generated lid-opening motion.
    For the dishwasher, the robot grasps the handle and follows the generated door motion.
    These examples illustrate how ArtiMo provides object-centric interaction targets that can be retargeted to downstream robotic execution.
    }
    \label{application}
    \vspace{-0.15in}

\end{figure*}

\fi\section{Application to Robotic Manipulation} \label{sec:robot_application} Beyond animation generation, we further investigate whether the structured motion plans produced by ArtiMo can serve as executable object-centric targets for downstream robotic manipulation. Manipulating an articulated object requires more than specifying where a part should move: the robot must identify \emph{which part should be interacted with}, determine \emph{how that interaction affects the object state}, and respect \emph{causal dependencies between successive motions}. These requirements closely match the information already inferred by ArtiMo during motion generation. We therefore reuse the output of our framework directly for robotic execution. Given an ArtiMo-generated motion plan, we identify the stages requiring external interaction and sample the corresponding trajectories of the manipulated object links. A feasible contact configuration is grounded on the relevant part, and the resulting object-space trajectory is retargeted to the robot end-effector and executed through inverse kinematics. Thus, the robot trajectory is not independently authored for each example, but is derived from the object motion and interaction sequence inferred by ArtiMo. This object-centric representation is particularly useful when the desired motion alone is insufficient to determine the correct interaction. For example, opening a door is not represented simply as rotating the door panel. ArtiMo identifies the handle as the control part and plans its actuation before the subsequent door motion, allowing the robot to first manipulate the handle and then follow the generated door-opening trajectory. Similarly, directly lifting the kettle lid could reproduce its final geometric motion while bypassing the actual operating mechanism. Instead, ArtiMo identifies the release button as the control and models the lid opening as its downstream effect; consequently, the robot presses the button rather than directly manipulating the lid. The dishwasher example further demonstrates direct grasp-and-follow retargeting, where the generated door trajectory provides the motion target after contact is established. As shown in Fig.~\ref{application}, these examples illustrate an important distinction between a standalone trajectory and the structured output of ArtiMo. A trajectory primarily specifies \emph{where} an articulated part moves, whereas ArtiMo additionally represents \emph{which interaction causes the motion}, \emph{which part should be manipulated}, and \emph{how multiple articulated states are causally ordered}. This enables the same object-centric motion representation to support different robotic interaction patterns, including direct grasp-and-follow manipulation and contact-triggered articulated motion. These results suggest that ArtiMo can serve as an intermediate representation between language-level intent and embodiment-specific actions, and may further provide structured motion and interaction supervision for downstream embodied or VLA-style policies.

\section{Conclusion}

We presented ArtiMo, an agent-driven framework for
action-conditioned articulated mesh animation. Given a URDF file,
mesh geometry, and a natural-language action prompt, ArtiMo combines
articulation priors with multimodal semantic reasoning to infer
action-relevant links, causal dependencies, and executable motion
plans. To improve robustness beyond single-pass generation, we further
introduced a critic--actor refinement loop that diagnoses motion errors
from rendered visual feedback and updates the animation plan
accordingly. We also constructed a benchmark of 225 articulated
animations covering causal and non-causal actions, together with
part-normalized 3D and 2D evaluation protocols designed to capture
small but causally important part movements. Experiments show that
ArtiMo outperforms generic 3D and 2D animation baselines in both
quantitative metrics and qualitative comparisons, while ablations
confirm the importance of structured planning and iterative
refinement. Beyond animation itself, we further demonstrate that the resulting object-centric motion plans can be retargeted to robotic execution. Because ArtiMo explicitly represents manipulated parts, interaction-induced effects, and causal motion ordering, its outputs provide more than geometric trajectories and can act as executable intermediate targets for embodiment-specific actions. We believe this broader connection between language-level intent, causal articulated motion, and downstream execution provides a promising direction toward more general and reusable representations of articulated-object behavior.

% We believe this work provides a step toward practical
% action-driven 4D animation generation for articulated objects,
% especially in settings where mechanically coupled, state-dependent,
% and environment-mediated motions are essential.
\clearpage
\bibliographystyle{ieeenat_fullname}\bibliography{refer_updated_v2}

\clearpage
\input{supplementary}

\end{document}

%% file: supplementary.tex
\setcounter{section}{0}
\setcounter{subsection}{0}
\setcounter{figure}{0}
\setcounter{table}{0}
\setcounter{equation}{0}
\renewcommand{\thesection}{\arabic{section}}
\renewcommand{\thesubsection}{\thesection.\arabic{subsection}}
\renewcommand{\thefigure}{\arabic{figure}}
\renewcommand{\thetable}{\arabic{table}}
\renewcommand{\theequation}{\arabic{equation}}

\twocolumn[
\begin{@twocolumnfalse}
\begin{center}
    {\LARGE\bfseries ArtiMo: Agent-Driven Articulated Mesh Animation\\[0.35em]}
    {\Large Supplementary Material}
\end{center}
\vspace{0.5em}
\end{@twocolumnfalse}
]

\section{Preliminaries and Input Representation}
\label{sec:supp_input_representation}

\subsection{URDF Representation and Gap}
\label{URDF}
The Unified Robot Description Format (URDF) is a widely used robot description format in the ROS ecosystem~\cite{quigley2009ros}. A URDF file represents an articulated object as a kinematic tree or graph composed of rigid links and joints. Links correspond to rigid parts of the object and may contain visual geometry, collision geometry, and inertial properties. Joints define the kinematic relationship between a parent link and a child link, including the joint type, motion axis, joint origin, and optional motion limits~\cite{tola2024understanding}. Common joint types include revolute joints, which rotate within a bounded angular range; continuous joints, which rotate without angular limits; prismatic joints, which translate along an axis; and fixed joints, which attach two links without relative motion.

In our setting, URDF provides the low-level executable structure for animation. The joint axis determines the local direction of motion, the joint origin defines the pivot or sliding frame, and the joint limits define the feasible range of motion. Together with the mesh geometry associated with each link, these fields make it possible to convert a motion plan into time-varying link poses. This makes URDF a compact and executable representation for articulated objects, and it is widely used in simulation environments and articulated-object datasets such as PartNet-Mobility~\cite{xiang2020sapien}.

Although URDF provides strong kinematic priors, it does not specify how an object should move in response to a natural-language action. In other words, URDF describes the feasible motion space of an articulated object, but not the action-conditioned motion trajectory. It does not indicate which link is the user-manipulated control, which link is the downstream effect, whether two joints are causally coupled, or how multiple phases should be temporally ordered. For example, a URDF file can specify that a trash bin contains a pedal joint and a lid joint, together with their axes and limits. However, it does not state that pressing the pedal should first actuate the pedal joint, then trigger the lid joint after a short causal delay, and finally hold the lid open if the desired action is ``fully open the trash bin.'' Similarly, URDF can encode that a door handle and a door panel are separate joints, but it does not specify whether the handle should self-return, whether the door should remain open, or how the handle and door motions should overlap in time.

This gap is central to our task. Given an action prompt, the system must infer which link is directly manipulated, which links are affected, which joints should move, what target states should be reached, and how the motions should be temporally coordinated. These requirements go beyond static articulation description and require semantic grounding, causal reasoning, and executable motion planning. Our framework therefore treats URDF as a low-level articulation prior and uses multimodal agentic reasoning to convert the action prompt into a time-parameterized animation plan.

\subsection{Reference and Overlay Images}

\begin{figure}[t]
    \centering
    \includegraphics[width=\columnwidth]{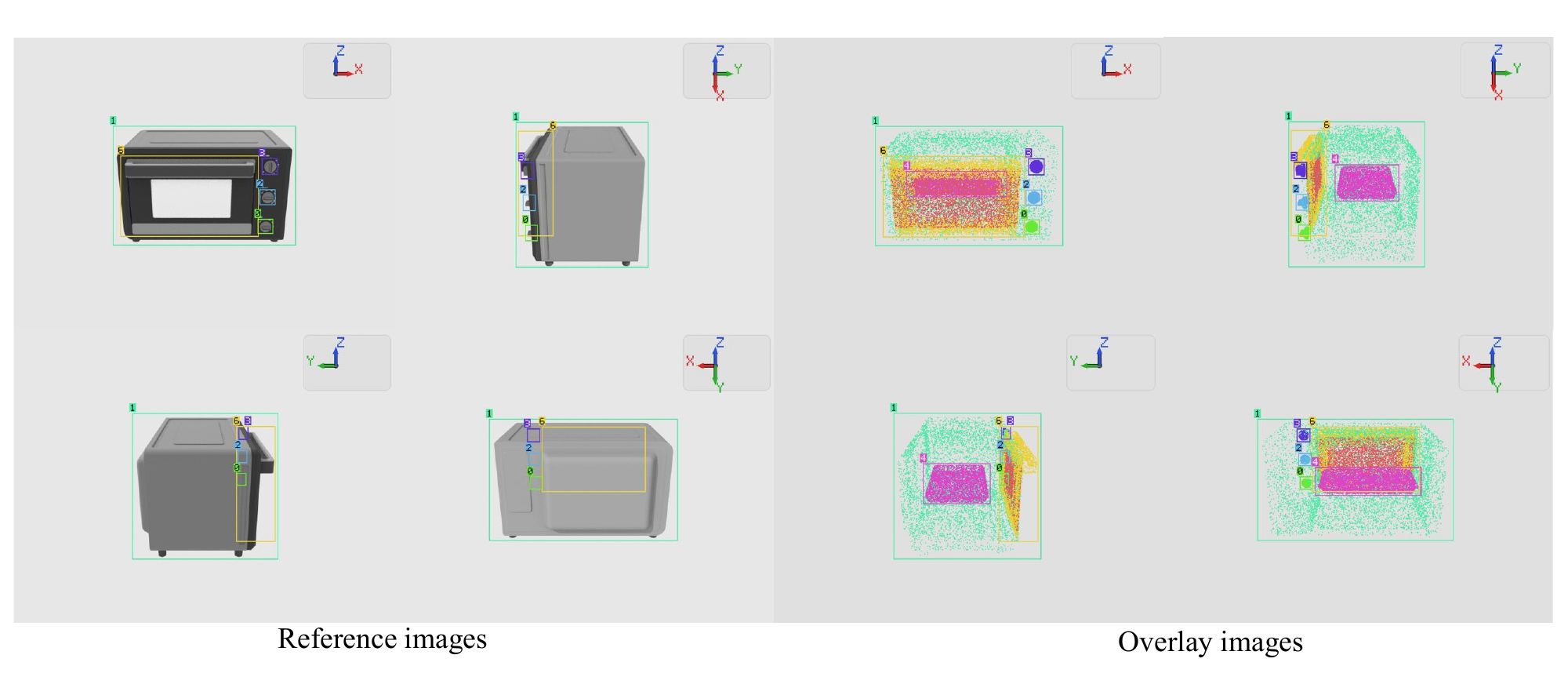}
    \caption{Example overlay and reference images used for link grounding.}
    \label{fig:supp_overlay_reference}
\end{figure}

We present example reference and overlay images in Fig.~\ref{fig:supp_overlay_reference}. Reference images show the assembled object appearance, but they may not expose internal or occluded articulated structures. Overlay images complement them by assigning different colors and compact labels to different links, revealing both visible and internal parts and indicating the spatial extent of each link. Together, reference and overlay views provide the VLM with the visual evidence needed to reason about link semantics, affordances, and causal relations.

\begin{figure}[!t]
    \centering
    \includegraphics[width=\linewidth]{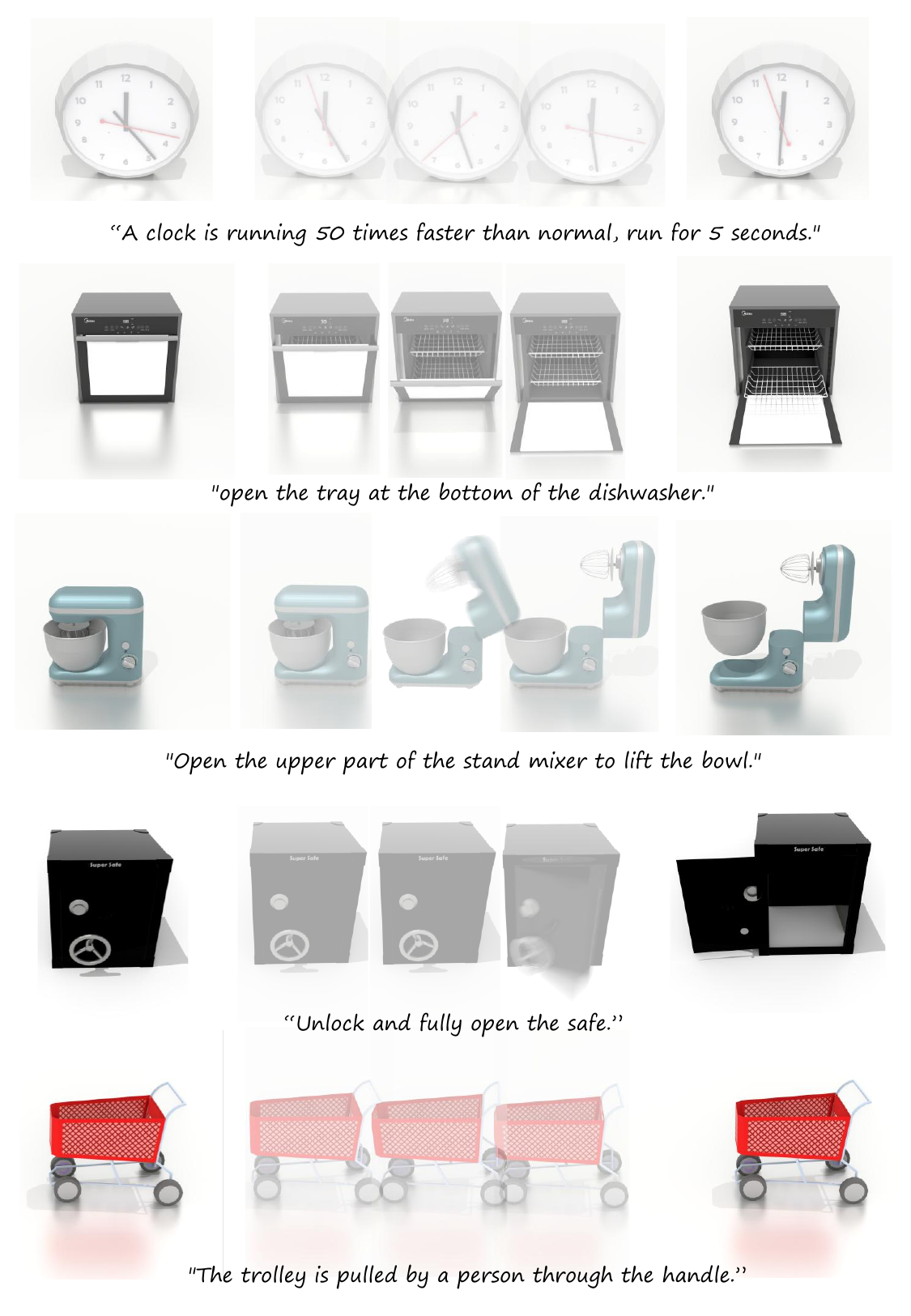}
    \caption{More qualitative results from ArtiMo.}
    \label{fig:placeholder}
\end{figure}

\input{supplementary_text/input_representation}

\FloatBarrier

\input{supplementary_text/prompt}

\section{Benchmark Construction and Evaluation Protocol}
\label{sec:supp_benchmark}

\input{benchmark_details}
\input{benchmark}

\section{Additional Experiments}
\label{sec:supp_additional_experiments}

\input{supplementary_text/additional_experiments}

\section{Limitations}
\label{sec:supp_failure}

Our method depends on the quality of the input articulation structure. Incorrect joint axes, origins, or link bindings can lead to implausible motion even when the inferred causal plan is correct. The system may also struggle with very small, textureless, or occluded parts that are difficult to ground from rendered views. In addition, our current motion plans emphasize kinematic and causal correctness rather than full physical simulation, so parameters for spring return, damping, or rolling are only approximate. Finally, since many action prompts allow multiple valid timings, our benchmark focuses on ordered key states rather than exact frame-level trajectories. Future work may combine stronger articulation reconstruction, physics-based validation, and learned motion critics to further improve robustness.

%% file: supplementary_text/input_representation.tex
\subsection{Joint Summary and Scale Context}
\label{sec:supp_joint_scale}

For each asset, the VLM receives a text summary of the URDF joint graph. Each record exposes the joint type, parent--child relation, axis, feasible limits, and local origin, grounding semantic reasoning in executable kinematics rather than rendered appearance alone.

\begin{schemabox}{URDF Joint Summary Schema}
\{\par
\quad "joint\_name": "joint\_id",\par
\quad "type": "revolute|prismatic|fixed",\par
\quad "parent": "link\_A", "child": "link\_B",\par
\quad "axis": [x, y, z],\par
\quad "limit": \{"lower": L, "upper": U,\par
\quad\quad\quad\quad\quad "effort": E, "velocity": V\},\par
\quad "origin": \{"xyz": [x, y, z],\par
\quad\quad\quad\quad\quad "rpy": [roll, pitch, yaw]\}\par
\}
\end{schemabox}

We additionally derive a scale context from the URDF and the reference GLB. Object and link bounding-box extents calibrate plausible motion magnitude; the object diagonal and GLB-to-URDF ratio calibrate camera distance and conversion between the rendered mesh and the kinematic frame. These quantities are numerical inputs to planning and rendering, not learned labels.

\begin{schemabox}{Scale Context Schema}
\{\par
\quad "asset": "asset\_id",\par
\quad "unit\_assumption": "meters\_like\_urdf\_units",\par
\quad "object\_bbox\_extents\_m": [dx, dy, dz],\par
\quad "object\_diag\_m": d,\par
\quad "link\_bbox\_extents\_m": \{"link\_X": [dx, dy, dz]\},\par
\quad "joint\_child\_link\_bbox\_extents\_m": \{...\},\par
\quad "median\_revolute\_child\_radius\_m\_est": r,\par
\quad "reference\_glb\_diag\_m": $d_{\mathrm{glb}}$,\par
\quad "glb\_to\_urdf\_scale\_ratio": s\par
\}
\end{schemabox}

%% file: supplementary_text/prompt.tex
\section{Agentic Animation Pipeline Details}
\label{sec:supp_pipeline}

We summarize the essential contract of each shared prompt below. Asset-specific
fields are filled at runtime..

\subsection{Animation Perceiver}
\label{perceiver}
The Perceiver grounds the user action to canonical URDF links and separates the
directly manipulated control from its downstream effects.

\begin{promptlisting}{Animation Perceiver Prompt Template (Low-Level Contract)}
INPUTS
  USER_ACTION_TEXT; LINK_INVENTORY; IMAGE_LABEL_MAPPING;
  OVERLAY_AND_REFERENCE_IMAGES; SCALE_CONTEXT_JSON;
  URDF_JOINT_SUMMARY; optional MASK_* images and conditioning text.
OUTPUT: STRICT JSON only; no prose/Markdown.

GROUNDING / COORDINATES
  - semantics.links must cover every visible link. "name" must be copied
    from LINK_INVENTORY; do not invent or normalize a link identifier.
  - Overlay labels and boxes are authoritative for identity/localization;
    reference renders are used only for appearance and affordance cues.
  - World legend: +X=red, +Y=green, +Z=blue; +Z is up. Never convert an
    action direction from image left/right.
  - A MASK_* attachment is a target mask, not a reference view. Map each
    masked region to a canonical link before using text to infer intent.
  - Use scale context to judge motion magnitude relative to the
  object and link size.

TARGET / EFFECT RESOLUTION
  - target_link is the first physically manipulated link in the causal
    chain, not automatically the largest or final moving link.
  - Priority: explicitly named control > visible/plausible dedicated
    control (button/latch/pedal/knob/handle) > directly moved part.
  - For multiple masked targets, emit all canonical names in target_links
    and keep the primary one in target_link.
  - target_role is exactly "control" or "direct_object".
  - Put downstream consequence links in effects.effect_links. Every joint
    named by a coupling rule must also occur in effects.joint_targets.
  - If there is no causal effect, use has_causal=false and set action,
    effects, and causal_segments to null. This is invalid for a requested
    continuous/speed-change motion unless the text explicitly denies motion.

NUMERIC / TEMPORAL ENCODING
  - Whole-object push/pull/drag/roll requires direction_axis_world, chosen from {[1,0,0],[-1,0,0],[0,1,0],[0,-1,0],[0,0,1],[0,0,-1]}.
  - Encode articulated targets per joint as upper_limit, lower_limit,
    alpha*upper_limit, alpha*lower_limit, or velocity:<number>. Never emit
    velocity:fast, velocity:unknown, or another symbolic speed.
  - control_return_behavior is self_return | stays | unknown. A return is
    allowed only if it does not undo the requested final state; otherwise
    use stays unless release/return/close is explicitly requested.
  - Use causal_segments only for causally distinct phases. Logical order
    does not force non-overlap: an effect may start after the causal delay
    while the control remains engaged.

OUTPUT SCHEMA
{
  "semantics": {"links": [
    {"name":"link_X","label":"...","affordance":["..."],
     "conf":0.0,"description":"..."}
  ]},
  "causal": {
    "has_causal": true,
    "action": {
      "primitive":"...","target_link":"link_X",
      "target_links":["link_X"],"target_role":"control",
      "magnitude":0.0,"direction_axis_world":[1,0,0],
      "control_return_behavior":"self_return|stays|unknown"
    },
    "effects": {
      "effect_links":["link_Y"],
      "joint_targets":[{"joint":"joint_Y","to":"upper_limit"}],
      "modes":[{"name":"...","set":true}],
      "coupling_rules":["..."]
    }
  },
  "causal_segments": null
}

SEGMENT FORM (only when needed)
{"segment_id":"S1","time_hint":{"order_index":0,"overlap_ok":true},
 "action":{...},"effects":{...}}
\end{promptlisting}

\subsection{Animation Planner}
\label{animation planner}
The Planner converts the Perceiver JSON and URDF limits into an executable
joint-level timeline.

\begin{promptlisting}{Animation Planner Prompt Template (Low-Level Contract)}
INPUTS
  USER_ACTION_TEXT; SCALE_CONTEXT_JSON; numeric URDF_JOINT_SUMMARY;
  Animation Perceiver causal JSON.
OUTPUT: STRICT JSON only. 

PERCEIVER AUTHORITY
  - Preserve target/effect identities, causal order, direction_axis_world,
    and each explicit joint target exactly. Never mirror paired joints,
    swap upper/lower limits, or force symmetric signs/magnitudes.
  - velocity:<number> fixes the signed omega target. For
    alpha*upper_limit / alpha*lower_limit, preserve q_target_expr and also
    resolve q_target_rad from that joint's numeric URDF limit.
  - Use only joints present in effects.joint_targets (top level or segment);
    one articulated control must contain exactly one "joint" field.

TIMELINE HARD RULES
  - phase_type is one of control_actuation, causal_latency,
    control_release, effect_motion, settle, hold.
  - A control-triggered effect uses at least:
      control_actuation -> causal_latency_or_release -> effect_motion.
    Do not merge control and main effect motion unless rigid coupling is
    explicit.
  - Enforce effect_start_time >= control_onset_time + delay_min:
      0.05 s button/latch/switch release;
      0.10 s spring-loaded lid/door;
      0.00 s only for direct grasped-part motion.
    effect_start_time may precede control_end_time (logical overlap).
  - Every segment has numeric t0,t1 with 0 <= t0 < t1 <= duration_s.
    fps in {24,30}; duration_s in [1,8].

CONTROL RECORDS (no unlisted fields; never "joints":[...])
  base_velocity:
    {mode,axis_world,v_mps}
  base_velocity_decay:
    {mode,axis_world,v0_mps,tau_s}
  joint_velocity:
    {mode,joint,omega_radps,ramp_to_omega_radps,decay}
  joint_position:
    {mode,joint,q_target_expr OR q_target_rad,q_start_rad,curve}
  hold_position:
    {mode,joint}
  spring_return:
    {mode,joint,spring_k,damping_c,rest_position}
  mode_set:
    {mode,name,set}

  omega_radps in [-20,20]; spring_k in [0,20]; damping_c in [0,5].
  Use base_* only for whole-object motion and copy direction_axis_world
  verbatim. Use joint_velocity for continuous/cyclic rotation.
  A decay.min_omega_radps is an unsigned magnitude floor; the sign remains
  in omega_radps or ramp_to_omega_radps.

RETURN / HOLD
  - spring_return is allowed only for a real restoring mechanism. Do not
    create a release phase merely because the action ends.
  - If release would undo the requested terminal state, hold the sustaining
    control through effect_motion/hold unless release was requested.
  - A transient latch/handle may return while the door/lid remains open;
    allocate enough return time before increasing stiffness or damping.

OUTPUT SCHEMA
{
  "meta":{"fps":30,"duration_s":0.0},
  "physics":{},
  "timeline":[{
    "name":"...","phase_type":"effect_motion",
    "t0":0.0,"t1":0.0,
    "controls":[
      {"mode":"joint_position","joint":"joint_X",
       "q_start_rad":0.0,"q_target_expr":"upper_limit",
       "q_target_rad":0.0,"curve":"ease_out"}
    ]
  }],
  "timing_checks":{
    "control_onset_time":0.0,"control_peak_time":0.0,
    "effect_start_time":0.0,"enforced_delay_s":0.0
  }
}
\end{promptlisting}

\subsection{Coverage-View Selection}
\label{sec:supp_coverage_schema}
The initial camera bank contains four object-centered views at
azimuths $0^\circ$, $90^\circ$, $180^\circ$, and $270^\circ$, each
with an elevation of $20^\circ$, a unit distance scale, and a
$35^\circ$ field of view. Additional views are selected from the
discrete camera parameter bank when the canonical views provide
insufficient visibility or motion observability.

The selector chooses a diagnostic view independently for each expected moving
link and requests additional camera views only when coverage is insufficient.

\begin{promptlisting}{Coverage-View Selection VLM Prompt Template (Low-Level Contract)}
INPUTS
  Separate current textured renders; CURRENT_VIEWSPECS_JSON;
  COVERAGE_REPORT_JSON; CAUSAL_SEMANTIC_JSON; EXPECTED_MOTION_JSON;
  optional MASK_* target attachments.
OUTPUT: STRICT JSON only. This module may rank views but must not edit the
action, causal graph, or motion plan.

PER-LINK SELECTION
  - selected_views_by_link must assign one CURRENT rendered view to each
    required link whenever possible; different links may use different
    views. selected_views is their deduplicated union.
  - A current view is usable iff the required link has BOTH a visible bbox
    and readable label text. Never infer identity from box position alone.
  - Rank, in order: exposed outer/surface-facing side; low occlusion; large
    readable link evidence; observable projected trajectory; acceptable
    distance. Reject back-facing, clipped, ambiguous, too-tight, or
    excessively distant alternatives.
  - For every required link, record bbox_visible, label_readable,
    trajectory_observability, evidence, chosen/rejected-view reasons, and
    why distance_scale is neither too near nor too far.

HIDDEN LINKS / NEW VIEWS
  - Never drop an initially hidden internal link. Identify the mover that
    exposes it and propose a camera facing the expected post-motion opening.
  - Any unassigned required link implies need_more_views=true.
  - If no current view can be selected confidently, selected_views must be
    [] and proposed_views must contain exactly four cameras.
  - Otherwise propose at most four new cameras, or [] when current coverage
    is sufficient.
  - Every proposed camera value must be copied from the allowed discrete
    sets: AZIMUTH_SET, ELEVATION_SET, DISTANCE_SCALE_SET, FOV_SET.

OUTPUT SCHEMA
{
  "selected_views_by_link":{
    "link_X":{"azimuth_deg":0,"elevation_deg":20,
              "distance_scale":1.0,"fov_deg":35}
  },
  "selected_views":[
    {"azimuth_deg":0,"elevation_deg":20,
     "distance_scale":1.0,"fov_deg":35}
  ],
  "need_more_views":true,
  "proposed_views":[
    {"azimuth_deg":90,"elevation_deg":30,
     "distance_scale":1.2,"fov_deg":35}
  ],
  "reason":"...",
  "selection_explanation":{
    "why_this_view":"...",
    "trajectory_reasoning":"...",
    "link_checks":[{
      "link":"link_X",
      "expected_motion":"rotation|translation|mixed|unknown",
      "bbox_visible":true,
      "label_readable":true,
      "trajectory_observability":"high|medium|low|unknown",
      "evidence":"...",
      "selection_reason":"...",
      "rejection_reason":"...",
      "distance_reason":"..."
    }]
  }
}
\end{promptlisting}

\subsection{Visual Critic and Deterministic Repair}
\label{diagnose}

\paragraph{Motion-cue construction.}
For each link--segment query, we use a head keyframe at the beginning
of the segment for identity grounding and a tail keyframe near its
endpoint for motion diagnosis. We sample one surface point near the
visible boundary of the target link, propagate it through the
executable forward-kinematics trajectory, and render its image-plane
track as a point-trajectory cue. A hollow circle marks the starting
position, while a filled circle marks the ending position. Straight
and circular arrows indicate projected translation and rotation,
respectively, while dot-out and cross-in badges indicate whether the
signed motion axis projects out of or into the camera plane.

Based on this compact motion representation, the VLM follows a structured diagnostic schema containing eight observation fields that summarize motion direction, view-dependent projection, magnitude, unintended motion, global motion, and release--return behavior. Based on these observations, the critic predicts issue labels drawn from five categories and produces a repair hint consisting of one of four
operator types, one of four direction values, and one of four discrete strength levels. This compact output space converts visual diagnosis into bounded and deterministic updates to the motion plan. For example, when a rice-cooker button fails to return after being pressed, the critic may refer to motion magnitude and release--return behavior to output
\texttt{NO\_RELEASE\_RETURN} with
\texttt{adjust\_timing} and \texttt{extra\_large}, causing the actor to
double the corresponding time window and provide sufficient time for
the return phase.
\begin{promptlisting}{Motion-Diagnosis VLM Prompt Template (Low-Level Contract)}
INPUTS
  Exactly one link/joint in one segment; HEAD and TAIL images;
  timeline_sample_catalog; trajectory_summary; plan_summary;
  scale_context; optional masks and previous reports
OUTPUT: STRICT JSON only.

SAMPLE / SOURCE ALIGNMENT
  1. Match TIMELINE_SAMPLE_n:<filename> exactly to catalog.file_name.
  2. Confirm segment_index/name, phase_type, link, joint, and
     single_link_trace before diagnosing.
  3. Identity comes from the HEAD bbox/label; motion is read from TAIL.
  4. Authoritative order:
       trajectory_summary.local_motion;
       local_motion_current_view.axis_projection_note/tag/projection;
       readable signed-axis or DOT/CROSS badge;
       sparse point tracks;
       drawn arrow shape (lowest).

SIGNED MOTION CONVERSION
  - Legend: +X red, +Y green, +Z blue. The named signed axis is required.
  - Rotation: DOT OUT keeps axis-relative CW/CCW; CROSS IN flips it.
    If current_view_direction exists, it is the final authoritative screen
    direction. A conflicting circular arrow is only ambiguous visual evidence.
  - Prismatic: project the named signed axis into the camera; do not apply
    DOT/CROSS conversion. Record Conversion as "none/not applicable".

INDEPENDENT CHECKS
  - Direction: compare converted signed motion with intended joint motion.
  - Magnitude/speed: use numeric start_q, end_q, delta_q,
    normalized_delta, joint limits, timeline controls, and scale context;
    never use apparent arrow length as the main magnitude measurement.
  - Extra motion: test whether this link should be static in this segment.
  - Return: use rest/target_return, final_q, remaining_error, and available
    return time; do not infer return from a single image.
  - Global/base motion is assessed separately from the local joint.

FIX VOCABULARY / SCOPE
  issue.code is exactly one of:
    WRONG_DIRECTION | UNEXPECTED_EXTRA_MOTION | EXCESSIVE_MOTION |
    MOTION_TOO_SMALL | NO_RELEASE_RETURN
  param_fix_hints.type is exactly:
    adjust_joint_target | adjust_joint_velocity |
    adjust_timing | adjust_direction
  direction is increase | decrease | flip | zero.
  strength is small | medium | large | extra_large.
  Every issue/hint must repeat the current joint, segment_index, and
  phase_type. The repair is parameter-only: do not change URDF, control
  mode/type, target/effect identity, or phase graph.

DETAILED REASONING FORMAT (all tags required)
  [Trajectory] axis_label=...; axis-relative direction=...
  [Projection] DOT OUT/CROSS IN or projected signed axis=...
  [Apparent] arrow read from image=...
  [Conversion] authoritative current-view direction=...
  [Magnitude] numeric actual-vs-intended judgment=...
  [ExtraMotion] unintended-motion judgment=...
  [Global] base direction or none/not applicable=...
  [Return] expected return or none/not applicable=...

OUTPUT SCHEMA
{
  "semantic_ok":true,"visibility_ok":true,
  "link_motion_type":
    "rotational_cw|rotational_ccw|prismatic|static_or_unclear",
  "detailed_reasoning":"[Trajectory] ...\n[Projection] ...\n...",
  "repairability":{
    "param_fixable":true,"structural_fix_needed":false,
    "preferred_repair_level":"param"
  },
  "issues":[{
    "code":"WRONG_DIRECTION","expected":"...","observed":"...",
    "segment_index":0,"phase_type":"effect_motion",
    "joint":"joint_X","link":"link_X"
  }],
  "param_fix_hints":[{
    "type":"adjust_direction","joint":"joint_X","segment_index":0,
    "phase_type":"effect_motion","direction":"flip","why":"..."
  }],
  "proposed_param_patch":null,
  "confidence":0.0
}
\end{promptlisting}

We allow at most three critic--actor repair rounds after the initial execution.

%% file: benchmark_details.tex
\subsection{Benchmark Sources and Statistics}
\label{sec:supp_benchmark_construction}

Our benchmark contains 180 unique source assets and 225 action instances collected from ArtVIP~\cite{jin2026artvip}, LightWheel~\cite{lightwheel_simready_2025}, and PartNet-Mobility~\cite{xiang2020sapien}. The resulting benchmark scale is comparable to ArtVIP V1\cite{jin2026artvip} and LightWheel\cite{lightwheel_simready_2025}, each with ~200 articulated assets. The source distribution is summarized in Table~\ref{tab:supp_benchmark_sources}, and the distribution across the 21 release-manifest object categories is shown in Table~\ref{tab:supp_category_coverage}.

\begin{table}[t]
\centering
\small
\setlength{\tabcolsep}{7pt}
\begin{tabular}{lr}
\toprule
Source & \# unique assets \\
\midrule
ArtVIP & 54 \\
LightWheel & 58 \\
PartNet-Mobility & 68 \\
\midrule
Total & 180 \\
\bottomrule
\end{tabular}
\caption{Benchmark sources. The release contains 225 action instances over these 180 unique assets.}
\label{tab:supp_benchmark_sources}
\end{table}

% For consistency with the main paper, all statistics use the 21 source object categories in the release manifest, rather than merging categories for compact reporting. The causal and non-causal dishwasher and microwave subsets are retained as distinct source categories because they have separate action annotations. Their asset coverage is listed in Table~\ref{tab:supp_category_coverage}.

\begin{table}[t]
\centering
\scriptsize
\setlength{\tabcolsep}{4pt}
\begin{tabular}{lr}
\toprule
Category & \# assets \\
\midrule
Basket & 3 \\
Bin & 6 \\
Clock & 12 \\
Dishwasher (causal) & 4 \\
Dishwasher (non-causal) & 16 \\
Door & 7 \\
Kettle & 13 \\
Laptop & 16 \\
Microwave (causal) & 2 \\
Microwave (non-causal) & 11 \\
Oven & 6 \\
Rice Cooker & 2 \\
Safe & 6 \\
Scissors & 14 \\
Single-tray Toaster Oven & 20 \\
Sliding Cabinet & 6 \\
Stand Mixer & 12 \\
Table & 2 \\
Toaster Oven & 6 \\
Trolley & 13 \\
Upper Cabinet & 3 \\
\midrule
Total & 180 \\
\bottomrule
\end{tabular}
\caption{Distribution of the 180 unique assets across the 21 release-manifest object categories. }
\label{tab:supp_category_coverage}
\end{table}

% Do not force the tall category table to finish before the following text.
% Let it float to the next column/page while Section 3.2 fills the remaining
% space; a barrier here creates a conspicuous blank area before Table 2.

%% file: benchmark.tex
\subsection{Benchmark Annotation Format}
\label{sec:supp_benchmark_format}

For each asset--action pair, expert annotators jointly inspect the
mesh, URDF structure, action instruction, structured annotation, and rendered animation. The annotation is revised and re-rendered until the intended joint motions, temporal relations, and terminal state are verified. Our benchmark represents each action as a structured annotation rather than a fixed frame-by-frame trajectory. This design is motivated by the fact that many articulated actions admit multiple valid timing realizations. For example, a door may open quickly or slowly, and a handle may return while the door is still opening or after it has already reached the target pose. Therefore, instead of enforcing a unique clock-time animation, we annotate the required motion as an ordered set of semantic phases, per-phase motion constraints, and optional temporal relations.

Each benchmark annotation contains three main components. The \texttt{phases} field defines the high-level action stages, such as \texttt{handle\_turn}, \texttt{door\_open}, or \texttt{hold\_end\_state}. The \texttt{constraints} field specifies what must happen in each phase, including the target joint or link, the required motion direction, and the target state such as joint angle, displacement, velocity, or return mode. The \texttt{relations} field specifies temporal dependencies between phases. Importantly, the phase order is a semantic ordering and does not by itself imply that two phases cannot overlap. Whether two phases are allowed to overlap is explicitly controlled by the corresponding relation.

A simplified annotation has the following structure:
\begin{promptlisting}{Benchmark Annotation Schema}
{
  "phases": [
    {
      "id": "phase_name",
      "order": 0,
      "duration_s": 0.0
    }
  ],
  "constraints": [
    {
      "phase": "phase_name",
      "subject": "joint_or_link_id",
      "constraints": {
        "motion": "motion_description",
        "target_q": 0.0,
        "displacement_m": 0.0,
        "omega_radps": 0.0,
        "return_mode": "self_return"
      }
    }
  ],
  "relations": [
    {
      "from_phase": "phase_A",
      "to_phase": "phase_B",
      "overlap_allowed": false,
      "start_rule": null
    }
  ]
}
\end{promptlisting}

\subsection{Non-Overlapping and Overlapping Motions}

\subsubsection{Non-overlapping phases}
Some actions require stages to happen sequentially. For example, when lifting a basket by its handles, the handles should first rotate upward, and the basket should then move upward as a whole. In this case, the relation between phases is marked with \texttt{overlap\_allowed=false}. This means that the second phase should be matched after the first phase has reached its target state.

\begin{promptlisting}{Example: Basket Lifting without Overlap}
{
  "phases": [
    {"id": "handles_raise", "order": 0, "duration_s": 0.6},
    {"id": "basket_lift", "order": 2, "duration_s": 1.0},
    {"id": "hold_end_state", "order": 3, "duration_s": 0.3}
  ],
  "constraints": [
    {
      "phase": "handles_raise",
      "subject": "joint_0",
      "constraints": {
        "motion": "rotate_about(+X,+)",
        "target_q": 2.094395102
      }
    },
    {
      "phase": "handles_raise",
      "subject": "joint_1",
      "constraints": {
        "motion": "rotate_about(+X,-)",
        "target_q": -2.094395102
      }
    },
    {
      "phase": "basket_lift",
      "subject": "base",
      "constraints": {
        "motion": "translate(+Z)",
        "displacement_m": 0.3
      }
    }
  ],
  "relations": [
    {
      "from_phase": "handles_raise",
      "to_phase": "basket_lift",
      "overlap_allowed": false
    },
    {
      "from_phase": "basket_lift",
      "to_phase": "hold_end_state",
      "overlap_allowed": false
    }
  ]
}
\end{promptlisting}

In this example, the two handle joints are first rotated upward, and the basket is lifted only after the handle-raising phase. Since both relations set \texttt{overlap\_allowed=false}, the evaluator treats these phases as sequential key states.

\subsubsection{Overlapping phases}
Other actions naturally contain concurrent motion. For instance, when opening a door with a knob or handle, the knob must first turn to unlatch the door, but the knob may begin returning to its rest pose while the door is still opening. Enforcing a strictly sequential order would be too rigid and would incorrectly penalize valid animations. We therefore allow certain relations to set \texttt{overlap\_allowed=true}. In this case, the \texttt{start\_rule} can specify that the target phase may begin after the source phase reaches a certain progress value, rather than after the source phase fully ends.

\begin{promptlisting}{Example: Door Opening with Overlap}
{
  "phases": [
    {"id": "knob_turn_unlatch", "order": 0, "duration_s": 0.3},
    {"id": "door_throw_open", "order": 1, "duration_s": 0.9},
    {"id": "knob_self_return", "order": 2, "duration_s": 1.8},
    {"id": "settle_open", "order": 3, "duration_s": 0.9}
  ],
  "constraints": [
    {
      "phase": "knob_turn_unlatch",
      "subject": "joint_1",
      "constraints": {
        "motion": "rotate_about(-Z,+)",
        "omega_radps": 8.0
      }
    },
    {
      "phase": "door_throw_open",
      "subject": "joint_2",
      "constraints": {
        "motion": "rotate_about(+Y,+)",
        "target_q": 1.5707963267948966
      }
    },
    {
      "phase": "knob_self_return",
      "subject": "joint_1",
      "constraints": {
        "return_mode": "self_return",
        "rest_position_q": 0.0
      }
    }
  ],
  "relations": [
    {
      "from_phase": "knob_turn_unlatch",
      "to_phase": "door_throw_open",
      "overlap_allowed": false
    },
    {
      "from_phase": "door_throw_open",
      "to_phase": "knob_self_return",
      "overlap_allowed": true,
      "start_rule": {
        "to_start_relative_to": "from_progress"
       }
    }
  ]
}
\end{promptlisting}

This example encodes two different temporal requirements. First, \texttt{door\_throw\_open} must occur after \texttt{knob\_turn\_unlatch}, because the door cannot open before it is unlatched. Second, \texttt{knob\_self\_return} is allowed to overlap with \texttt{door\_throw\_open}. The start rule indicates that the knob return may begin once the door-opening phase has started and reached a small amount of progress. This makes the annotation flexible enough to accept physically plausible variations, while still preserving the causal order that the knob must be actuated before the door opens.

\paragraph{Evaluation with overlap.}
During evaluation, non-overlapping phases are matched as ordered phase endpoints. For overlapping relations, the overlapped phase is not required to appear only after the previous phase endpoint. Instead, it is allowed to match within the valid interval implied by the relation. This avoids over-constraining natural motions such as handle return, button rebound, or pedal release. As a result, the benchmark evaluates whether the required causal states occur in a valid order, rather than enforcing a single rigid timeline.

%% file: supplementary_text/additional_experiments.tex
\input{supplementary_text/motion_type_results}

% \begin{table}[t]
% \centering
% \scriptsize
% \setlength{\tabcolsep}{5pt}
% \begin{tabular}{lrrr}
% \toprule
% Executions & Repairs & \# runs & Mean runtime (s) \\
% \midrule
% 1 & 0 & 95 & 212 \\
% 2 & 1 & 32 & 341 \\
% 3 & 2 & 24 & 437 \\
% 4 & 3 & 5 & 491 \\
% \bottomrule
% \end{tabular}
% \caption{Iterative behavior over 156 causal-pipeline runs.}
% \label{tab:supp_refinement_iterations}
% \end{table}

\subsection{Critic Diagnostic Accuracy}
\label{sec:supp_critic_accuracy}

As shown in Table~\ref{tab:supp_critic_injection}, we test the deployed critic on 44 cases spanning all four motion types and 13 object classes. Starting from a correct ground-truth plan, we inject one known local failure, regenerate the trajectory and sparse point-trajectory cues, and query the same critic used by the refinement loop. The critic receives the correct action/causal record and corrupted execution, but never the ground-truth animation. Detection denotes flagging the issue. The controlled test covers 168 injections over the the five injected-failure categories.

\begin{table}[t]
\centering
\scriptsize
\setlength{\tabcolsep}{3pt}
\begin{tabular}{lrrr}
\toprule
Injected failure & $N$ & Detection \\
\midrule
\texttt{WRONG\_DIRECTION} & 44 & 95.5\% \\
\texttt{MOTION\_TOO\_SMALL} & 44 & 95.5\% \\
\texttt{EXCESSIVE\_MOTION} & 24 & 91.7\% \\
\texttt{UNEXPECTED\_EXTRA\_MOTION} & 38 & 86.8\%\\
\texttt{NO\_RELEASE\_RETURN} & 18 & 100.0\% \\
\midrule
\textbf{Overall} & \textbf{168} & \textbf{93.5\%}  \\
\bottomrule
\end{tabular}
\caption{Critic performance on injected, visually verifiable local failures.}
\label{tab:supp_critic_injection}
\end{table}

\subsection{Runtime and Model Calls}
\label{sec:supp_runtime}

As shown in Table~\ref{tab:supp_runtime_calls}, runtime is measured per asset--action run over the all causal asset categories. The loop uses one initial LLM planner call, one initial VLM perception call, and VLM calls for coverage and motion diagnosis; deterministic parameter patches are not model calls. The measured mean is $282.2\pm133.6$ seconds, with 3.87 VLM calls and one LLM call per run. The initial pass averages 212 seconds, and subsequent refinements add roughly 95--130 seconds. Reconstructing the stored artifact timeline attributes about 44\% of tracked wall-clock time (112 seconds) to VLM API latency and 56\% (138 seconds) to rendering, geometry, trajectory, GLTF export, and the LLM call.

\begin{table}[t]
\centering
% \scriptsize
% \setlength{\tabcolsep}{4pt}
\begin{tabular}{lrrrr}
\toprule
Metric & Mean & Std & Min & Max \\
\midrule
Runtime (s) & 282.2 & 133.6 & 76 & 821 \\
VLM calls & 3.87 & 0.99 & 3 & 7 \\
LLM calls & 1.00 & 0.00 & 1 & 1 \\
Total model calls & 4.87 & 0.99 & 4 & 8 \\
\bottomrule
\end{tabular}
\caption{Runtime and model calls per causal-pipeline run.}
\label{tab:supp_runtime_calls}
\end{table}

\begin{table}[t]
\centering
\scriptsize
\setlength{\tabcolsep}{5pt}
\begin{tabular}{lccc}
\toprule
Run & P\_gIoU & P\_PC & P\_OccF1 \\
\midrule
1 & 0.954 & 0.894 & 0.665 \\
2 & 0.975 & 0.903 & 0.672 \\
3 & 0.972 & 0.912 & 0.690 \\
\midrule
Mean $\pm$ std & $0.967\pm0.011$ & $0.903\pm0.009$ & $0.676\pm0.013$ \\
\bottomrule
\end{tabular}
\caption{Independent three-run variance on the 32-case balanced subset.}
\label{tab:supp_variance}
\end{table}

\subsection{Multiple-Run Variance}
\label{sec:supp_variance}

As shown in Table~\ref{tab:supp_variance}, On a balanced 32-case subset (eight cases per motion type), we run the full agent three times end-to-end and evaluate all 96 outputs with the same 3D protocol. The standard deviation of the three run-level means is only 0.011 for P\_gIoU, 0.009 for P\_PC, and 0.013 for P\_OccF1. The mean per-case P\_gIoU standard deviation is 0.024, indicating that the headline conclusion is stable under foundation-model sampling variation.

%% file: supplementary_text/motion_type_results.tex
\subsection{Results by Motion Type}
\label{sec:supp_motion_type_results}

As shwon in Table \ref{tab:supp_motion_type_scores}, we assign each of the 225 cases to one of the four motion types using the ground-truth plan and annotation structure (control modes, phase/relation graph, joint coupling, and environmental transport): 71 independent, 58 mechanically coupled, 67 state-dependent, and 29 environment-mediated cases. 

\begin{table}[!t]
\centering
\textbf{ArtiMo (Full): mean $\pm$ std}\\[-2pt]
\tiny
\setlength{\tabcolsep}{3pt}
\renewcommand{\arraystretch}{1.08}
\begin{tabular}{lcccc}
\toprule
Type & $N$ & P\_gIoU & P\_PC & P\_OccF1 \\
\midrule
Independent & 71 & $0.976\pm0.201$ & $0.970\pm0.121$ & $0.903\pm0.206$ \\
Mech. coupled & 58 & $0.975\pm0.079$ & $0.945\pm0.131$ & $0.886\pm0.156$ \\
State-dependent & 67 & $0.998\pm0.013$ & $0.984\pm0.033$ & $0.950\pm0.100$ \\
Env.-mediated & 29 & $1.000\pm0.000$ & $0.953\pm0.040$ & $0.794\pm0.159$ \\
Overall & 225 & $0.985\pm0.120$ & $0.965\pm0.098$ & $0.899\pm0.167$ \\
\bottomrule
\end{tabular}
\caption{Performance by motion type. Ind., Mech., State, and Env. denote independent, mechanically coupled, state-dependent, and environment-mediated motion, respectively.}
\label{tab:supp_motion_type_scores}
\end{table}

%% file: refer_updated_v2.bib
@inproceedings{jiang2024animate3d,
  title     = {Animate3D: Animating Any 3D Model with Multi-view Video Diffusion},
  author    = {Jiang, Yanqin and Yu, Chaohui and Cao, Chenjie and Wang, Fan and Hu, Weiming and Gao, Jin},
  booktitle = {Advances in Neural Information Processing Systems},
  year      = {2024}
}

@inproceedings{wu2025animateanymesh,
  title     = {AnimateAnyMesh: A Feed-Forward 4D Foundation Model for Text-Driven Universal Mesh Animation},
  author    = {Wu, Zijie and Yu, Chaohui and Wang, Fan and Bai, Xiang},
  booktitle = {IEEE/CVF International Conference on Computer Vision},
  year      = {2025}
}

@inproceedings{uzolas2025motiondreamer,
  title     = {MotionDreamer: Exploring Semantic Video Diffusion Features for Zero-Shot 3D Mesh Animation},
  author    = {Uzolas, Lukas and Eisemann, Elmar and Kellnhofer, Petr},
  booktitle = {International Conference on 3D Vision},
  year      = {2025}
}

@inproceedings{li2025puppetmaster,
  title     = {Puppet-Master: Scaling Interactive Video Generation as a Motion Prior for Part-Level Dynamics},
  author    = {Li, Ruining and Zheng, Chuanxia and Rupprecht, Christian and Vedaldi, Andrea},
  booktitle = {IEEE/CVF International Conference on Computer Vision},
  year      = {2025}
}

@inproceedings{madaan2023selfrefine,
  title     = {Self-Refine: Iterative Refinement with Self-Feedback},
  author    = {Madaan, Aman and Tandon, Niket and Gupta, Prakhar and Hallinan, Skyler and Gao, Luyu and Wiegreffe, Sarah and Alon, Uri and Dziri, Nouha and Prabhumoye, Shrimai and Yang, Yiming and Gupta, Shashank and Majumder, Bodhisattwa Prasad and Hermann, Katherine and Welleck, Sean and Yazdanbakhsh, Amir and Clark, Peter},
  booktitle = {Advances in Neural Information Processing Systems},
  year      = {2023}
}

@inproceedings{yang2024holodeck,
  title     = {Holodeck: Language Guided Generation of 3D Embodied AI Environments},
  author    = {Yang, Yue and Sun, Fan-Yun and Weihs, Luca and VanderBilt, Eli and Herrasti, Alvaro and Han, Winson and Wu, Jiajun and Haber, Nick and Krishna, Ranjay and Liu, Lingjie and Callison-Burch, Chris and Yatskar, Mark and Kembhavi, Aniruddha and Clark, Christopher},
  booktitle = {IEEE/CVF Conference on Computer Vision and Pattern Recognition},
  year      = {2024}
}

@article{xia2026sage,
  title   = {SAGE: Scalable Agentic 3D Scene Generation for Embodied AI},
  author  = {Xia, Hongchi and Li, Xuan and Li, Zhaoshuo and Ma, Qianli and Xu, Jiashu and Liu, Ming-Yu and Cui, Yin and Lin, Tsung-Yi and Ma, Wei-Chiu and Wang, Shenlong and Song, Shuran and Wei, Fangyin},
  journal = {arXiv preprint arXiv:2602.10116},
  year    = {2026}
}

@inproceedings{feng2023layoutgpt,
  title     = {LayoutGPT: Compositional Visual Planning and Generation with Large Language Models},
  author    = {Feng, Weixi and Zhu, Wanrong and Fu, Tsu-Jui and Jampani, Varun and Akula, Arjun and He, Xuehai and Basu, Sugato and Wang, Xin Eric and Wang, William Yang},
  booktitle = {Advances in Neural Information Processing Systems},
  volume    = {36},
  year      = {2023}
}

@inproceedings{xiang2020sapien,
  title     = {{SAPIEN}: A SimulAted Part-Based Interactive ENvironment},
  author    = {Xiang, Fanbo and Qin, Yuzhe and Mo, Kaichun and Xia, Yikuan and Zhu, Hao and Liu, Fangchen and Liu, Minghua and Jiang, Hanxiao and Yuan, Yifu and Wang, He and Yi, Li and Chang, Angel X. and Guibas, Leonidas J. and Su, Hao},
  booktitle = {Proceedings of the IEEE/CVF Conference on Computer Vision and Pattern Recognition},
  pages     = {11097--11107},
  year      = {2020}
}

@inproceedings{jin2026artvip,
  title     = {{ArtVIP}: Articulated Digital Assets of Visual Realism, Modular Interaction, and Physical Fidelity for Robot Learning},
  author    = {Jin, Zhao and Che, Zhengping and Li, Tao and Zhao, Zhen and Wu, Kun and Zhang, Yuheng and Zhao, Yinuo and Liu, Zehui and Zhang, Qiang and Ju, Xiaozhu and Tian, Jing and Xue, Yousong and Tang, Jian},
  booktitle = {International Conference on Learning Representations},
  year      = {2026}
}

@article{li2025particulate,
  title   = {Particulate: Feed-Forward 3D Object Articulation},
  author  = {Li, Ruining and Yao, Yuxin and Zheng, Chuanxia and Rupprecht, Christian and Lasenby, Joan and Wu, Shangzhe and Vedaldi, Andrea},
  journal = {arXiv preprint arXiv:2512.11798},
  year    = {2025}
}

@inproceedings{quigley2009ros,
  title     = {{ROS}: An Open-Source Robot Operating System},
  author    = {Quigley, Morgan and Conley, Ken and Gerkey, Brian and Faust, Josh and Foote, Tully and Leibs, Jeremy and Wheeler, Rob and Ng, Andrew Y.},
  booktitle = {ICRA Workshop on Open Source Software},
  year      = {2009}
}

@article{tola2024understanding,
  title   = {Understanding {URDF}: A Dataset and Analysis},
  author  = {Tola, Daniella and Corke, Peter},
  journal = {IEEE Robotics and Automation Letters},
  year    = {2024},
  doi     = {10.1109/LRA.2024.3381482}
}

@inproceedings{rezatofighi2019giou,
  title={Generalized Intersection over Union: A Metric and a Loss for Bounding Box Regression},
  author={Rezatofighi, Hamid and Tsoi, Nathan and Gwak, JunYoung and Sadeghian, Amir and Reid, Ian and Savarese, Silvio},
  booktitle={CVPR},
  year={2019}
}

@inproceedings{fan2017pointset,
  title={A Point Set Generation Network for 3D Object Reconstruction from a Single Image},
  author={Fan, Haoqiang and Su, Hao and Guibas, Leonidas J.},
  booktitle={CVPR},
  year={2017}
}

@inproceedings{mescheder2019occupancy,
  title={Occupancy Networks: Learning 3D Reconstruction in Function Space},
  author={Mescheder, Lars and Oechsle, Michael and Niemeyer, Michael and Nowozin, Sebastian and Geiger, Andreas},
  booktitle={CVPR},
  year={2019}
}

@inproceedings{lin2014coco,
  title={Microsoft COCO: Common Objects in Context},
  author={Lin, Tsung-Yi and Maire, Michael and Belongie, Serge and Hays, James and Perona, Pietro and Ramanan, Deva and Doll{\'a}r, Piotr and Zitnick, C. Lawrence},
  booktitle={ECCV},
  year={2014}
}

@article{arbelaez2011contour,
  title={Contour Detection and Hierarchical Image Segmentation},
  author={Arbel{\'a}ez, Pablo and Maire, Michael and Fowlkes, Charless and Malik, Jitendra},
  journal={IEEE TPAMI},
  volume={33},
  number={5},
  pages={898--916},
  year={2011}
}

@article{borgefors1988chamfer,
  title={Hierarchical Chamfer Matching: A Parametric Edge Matching Algorithm},
  author={Borgefors, Gunilla},
  journal={IEEE TPAMI},
  volume={10},
  number={6},
  pages={849--865},
  year={1988}
}

@inproceedings{wang2019shape2motion,
  title={Shape2Motion: Joint Analysis of Motion Parts and Attributes from 3D Shapes},
  author={Wang, Xiaogang and Zhou, Bin and Shi, Yahao and Chen, Xiaowu and Zhao, Qinping and Xu, Kai},
  booktitle={IEEE/CVF Conference on Computer Vision and Pattern Recognition (CVPR)},
  year={2019}
}

@inproceedings{lei2023nap,
  title={NAP: Neural 3D Articulated Object Prior},
  author={Lei, Jiahui and Deng, Congyue and Shen, William B. and Guibas, Leonidas J. and Daniilidis, Kostas},
  booktitle={Advances in Neural Information Processing Systems (NeurIPS)},
  year={2023}
}

@inproceedings{chen2024urdformer,
  title={URDFormer: A Pipeline for Constructing Articulated Simulation Environments from Real-World Images},
  author={Chen, Zoey and Walsman, Aaron and Memmel, Marius and Mo, Kaichun and Fang, Alex and Vemuri, Karthikeya and Wu, Alan and Fox, Dieter and Gupta, Abhishek},
  booktitle={Robotics: Science and Systems (RSS)},
  year={2024}
}

@inproceedings{le2025articulate,
  title={Articulate-Anything: Automatic Modeling of Articulated Objects via a Vision-Language Foundation Model},
  author={Le, Long and Xie, Jason and Liang, William and Wang, Hung-Ju and Yang, Yue and Ma, Yecheng Jason and Vedder, Kyle and Krishna, Arjun and Jayaraman, Dinesh and Eaton, Eric},
  booktitle={International Conference on Learning Representations (ICLR)},
  year={2025}
}

@inproceedings{jiang2022ditto,
  title={Ditto: Building Digital Twins of Articulated Objects from Interaction},
  author={Jiang, Zhenyu and Hsu, Cheng-Chun and Zhu, Yuke},
  booktitle={IEEE/CVF Conference on Computer Vision and Pattern Recognition (CVPR)},
  year={2022}
}

@inproceedings{liu2023paris,
  title={PARIS: Part-level Reconstruction and Motion Analysis for Articulated Objects},
  author={Liu, Jiayi and Mahdavi-Amiri, Ali and Savva, Manolis},
  booktitle={IEEE/CVF International Conference on Computer Vision (ICCV)},
  year={2023}
}

@inproceedings{geng2023gapartnet,
  title={GAPartNet: Cross-Category Domain-Generalizable Object Perception and Manipulation via Generalizable and Actionable Parts},
  author={Geng, Haoran and Xu, Helin and Zhao, Chengyang and Xu, Chao and Yi, Li and Huang, Siyuan and Wang, He},
  booktitle={IEEE/CVF Conference on Computer Vision and Pattern Recognition (CVPR)},
  year={2023}
}

@inproceedings{mo2021where2act,
  title={Where2Act: From Pixels to Actions for Articulated 3D Objects},
  author={Mo, Kaichun and Guibas, Leonidas J. and Mukadam, Mustafa and Gupta, Abhinav and Tulsiani, Shubham},
  booktitle={IEEE/CVF International Conference on Computer Vision (ICCV)},
  year={2021}
}

@inproceedings{eisner2022flowbot3d,
  title={FlowBot3D: Learning 3D Articulation Flow to Manipulate Articulated Objects},
  author={Eisner, Ben and Zhang, Harry and Held, David},
  booktitle={Robotics: Science and Systems (RSS)},
  year={2022}
}

@inproceedings{wu2022vatmart,
  title={VAT-Mart: Learning Visual Action Trajectory Proposals for Manipulating 3D ARTiculated Objects},
  author={Wu, Ruihai and Zhao, Yan and Mo, Kaichun and Guo, Zhizheng and Wang, Yian and Wu, Tianhao and Fan, Qingnan and Chen, Xuelin and Guibas, Leonidas and Dong, Hao},
  booktitle={International Conference on Learning Representations (ICLR)},
  year={2022}
}

@article{witkin1988spacetime,
  title   = {Spacetime Constraints},
  author  = {Witkin, Andrew and Kass, Michael},
  journal = {ACM SIGGRAPH Computer Graphics},
  volume  = {22},
  number  = {4},
  pages   = {159--168},
  year    = {1988},
  doi     = {10.1145/378456.378507}
}

@article{perlin1995responsive,
  title   = {Real Time Responsive Animation with Personality},
  author  = {Perlin, Ken},
  journal = {IEEE Transactions on Visualization and Computer Graphics},
  volume  = {1},
  number  = {1},
  pages   = {5--15},
  year    = {1995},
  doi     = {10.1109/2945.468392}
}

@article{garrett2021tamp,
  title   = {Integrated Task and Motion Planning},
  author  = {Garrett, Caelan Reed and Chitnis, Rohan and Holladay, Rachel and Kim, Beomjoon and Silver, Tom and Kaelbling, Leslie Pack and Lozano-P{\'e}rez, Tom{\'a}s},
  journal = {Annual Review of Control, Robotics, and Autonomous Systems},
  volume  = {4},
  pages   = {265--293},
  year    = {2021},
  doi     = {10.1146/annurev-control-091420-084139}
}

@inproceedings{todorov2012mujoco,
  title     = {{MuJoCo}: A Physics Engine for Model-Based Control},
  author    = {Todorov, Emanuel and Erez, Tom and Tassa, Yuval},
  booktitle = {IEEE/RSJ International Conference on Intelligent Robots and Systems},
  pages     = {5026--5033},
  year      = {2012},
  doi       = {10.1109/IROS.2012.6386109}
}

@article{chen2025robotwin2,
  title   = {{RoboTwin} 2.0: A Scalable Data Generator and Benchmark with Strong Domain Randomization for Robust Bimanual Robotic Manipulation},
  author  = {Chen, Tianxing and Chen, Zanxin and Chen, Baijun and Cai, Zijian and Liu, Yibin and Li, Zixuan and Liang, Qiwei and Lin, Xianliang and Ge, Yiheng and Gu, Zhenyu and Deng, Weiliang and Guo, Yubin and Nian, Tian and Xie, Xuanbing and Chen, Qiangyu and Su, Kailun and Xu, Tianling and Liu, Guodong and Hu, Mengkang and Gao, Huan-ang and Wang, Kaixuan and Liang, Zhixuan and Qin, Yusen and Yang, Xiaokang and Luo, Ping and Mu, Yao},
  journal = {arXiv preprint arXiv:2506.18088},
  year    = {2025}
}

@inproceedings{gu2025blendergym,
  title     = {{BlenderGym}: Benchmarking Foundational Model Systems for Graphics Editing},
  author    = {Gu, Yunqi and Huang, Ian and Je, Jihyeon and Yang, Guandao and Guibas, Leonidas J.},
  booktitle = {IEEE/CVF Conference on Computer Vision and Pattern Recognition},
  pages     = {18574--18583},
  year      = {2025},
  doi       = {10.1109/CVPR52734.2025.01731}
}

@article{liu2025ir3dbench,
  title   = {{IR3D-Bench}: Evaluating Vision-Language Model Scene Understanding as Agentic Inverse Rendering},
  author  = {Liu, Parker and Li, Chenxin and Li, Zhengxin and Wu, Yipeng and Li, Wuyang and Yang, Zhiqin and Zhang, Zhenyuan and Lin, Yunlong and Han, Sirui and Feng, Brandon Y.},
  journal = {arXiv preprint arXiv:2506.23329},
  year    = {2025}
}

@article{yin2026viga,
  title   = {Vision-as-Inverse-Graphics Agent via Interleaved Multimodal Reasoning},
  author  = {Yin, Shaofeng and Ge, Jiaxin and Wang, Zora Zhiruo and Li, Xiuyu and Black, Michael J. and Darrell, Trevor and Kanazawa, Angjoo and Feng, Haiwen},
  journal = {arXiv preprint arXiv:2601.11109},
  year    = {2026}
}

@misc{lightwheel_simready_2025,
  title={Lightwheel Sim-Ready Assets: High-Quality USD Assets for NVIDIA Isaac Sim},
  author={{Lightwheel}},
  howpublished={GitHub repository},
  url={https://github.com/LightwheelAI/Lightwheel-simready-asset},
  year={2025},
  version={v1},
  note={259 robotics simulation assets under CC BY-NC 4.0}
}
